\documentclass{article}

\PassOptionsToPackage{square,sort,comma,numbers}{natbib}

\pdfoutput=1

\usepackage[final]{fl_neurips_2022}

\usepackage{graphicx}

\usepackage[utf8]{inputenc} %
\usepackage[T1]{fontenc}    %
\usepackage{hyperref}       %
\usepackage{url}            %
\usepackage{booktabs}       %
\usepackage{amsfonts}       %
\usepackage{nicefrac}       %
\usepackage{microtype}      %
\usepackage{xcolor}         %

\usepackage{subfig}
\usepackage{amsthm}
\usepackage{amsmath,amssymb,amsfonts}
\usepackage[ruled,noend]{algorithm2e}
\usepackage{multicol}
\usepackage{verbatim}
\usepackage{xfrac}
\usepackage{mathtools}
\usepackage{graphicx}
\usepackage{textcomp}

\usepackage{chngcntr}
\usepackage{apptools}
\newtheorem{theorem}{Theorem}
\newtheorem{assumption}{Assumption}
\AtAppendix{\counterwithin{assumption}{section}}

\newcommand{\E}{\mathbb{E}}

\title{Federated Progressive Sparsification (Purge-Merge-Tune)\textsuperscript{+}}

\author{%
  Dimitris Stripelis \\
  Information Sciences Institute\\
  University of Southern California\\
  Marina Del Rey, CA  90292\\
  \texttt{stripeli@isi.edu} \\
  \And
  Umang Gupta \\
  Information Sciences Institute\\
  University of Southern California\\
  Marina Del Rey, CA  90292\\
  \texttt{umanggup@isi.edu} \\
  \AND
  Greg Ver Steeg \\
  Information Sciences Institute\\
  University of Southern California\\
  Marina Del Rey, CA  90292\\
  \texttt{gregv@isi.edu} \\
  \And
  Jos\'{e} Luis Ambite \\
  Information Sciences Institute\\
  University of Southern California\\
  Marina Del Rey, CA  90292\\
  \texttt{ambite@isi.edu} \\
}

\begin{document}

\maketitle

\begin{abstract}
    Federated learning is a promising approach for training machine learning models on decentralized data while keeping data private at each client.  
    Model sparsification seeks to produce small neural models with comparable performance to large models; for example, for deployment on clients with limited memory or computational capabilites. We present \textit{FedSparsify}, a simple yet effective sparsification strategy for federated training of neural networks based on progressive weight magnitude pruning.  \textit{FedSparsify} learns subnetworks smaller than 10\% of the original network size with similar or better accuracy. Through extensive experiments, we demonstrate that \textit{FedSparsify} results in an average 15-fold model size reduction, 4-fold model inference speedup, and a 3-fold training communication cost improvement across various challenging domains and model architectures. Finally, we also theoretically analyze \textit{FedSparsify}'s impact on the convergence of federated training. Overall, our results show that \textit{FedSparsify} is an effective method to train extremely sparse and highly accurate models in federated learning settings.
\end{abstract}

\section{Introduction}

Federated Learning~\cite{mcmahan2017communication,li2020federated,yang2019federated} has emerged as the standard distributed machine learning paradigm to train neural networks without sharing data. Compared to traditional centralized machine learning approaches, which require data from different sources to be aggregated at a single location, Federated Learning allows private data to remain at its original location. Each data source (client) trains the model on its private data and sends only its locally-trained model parameters (e.g., gradients, weights) to a central server. 
Model sparsification, aka model pruning, (e.g.,~\cite{liu2018rethinking,frankle2018lottery}) seeks to produce small neural models with similar performance to large models.
In this work, we focus on learning extremely sparse and highly accurate federated models that can be deployed for fast and cost-efficient inference in resource-constrained devices~\cite{liu2022federated,hoefler2021sparsity}, while reducing associated communication and training costs.

Previous methods to speed up federated training and reduce model size include knowledge transfer~\cite{he2020group}, neural architecture search~\cite{xu2020federated}, and quantization~\cite{reisizadeh2020fedpaq,xu2020ternary}. Recently several methods have also been proposed to prune federated model weights, such as PruneFL~\cite{bibikar2022federated} and FedDST~\cite{bibikar2022federated}. These approaches are designed especially for federated training, are complicated, and often underperform non-pruning baselines.

We propose \textit{FedSparsify}, a simple iterative federated pruning procedure that progressively prunes model parameters based on weight magnitude at the end of each federation round. Our method simultaneously learns smaller neural networks for faster inference (and training) and reduces training communication costs by gradually decreasing the total number of model parameters exchanged between the clients and the server. Such iterative/progressive sparsification has been shown to lead to superior results in centralized settings~\cite{frankle2018lottery,liu2018rethinking}, but has been largely unexplored in federated settings.

\begin{figure}[htpb]
    \centering
    \includegraphics[width=.8\textwidth]{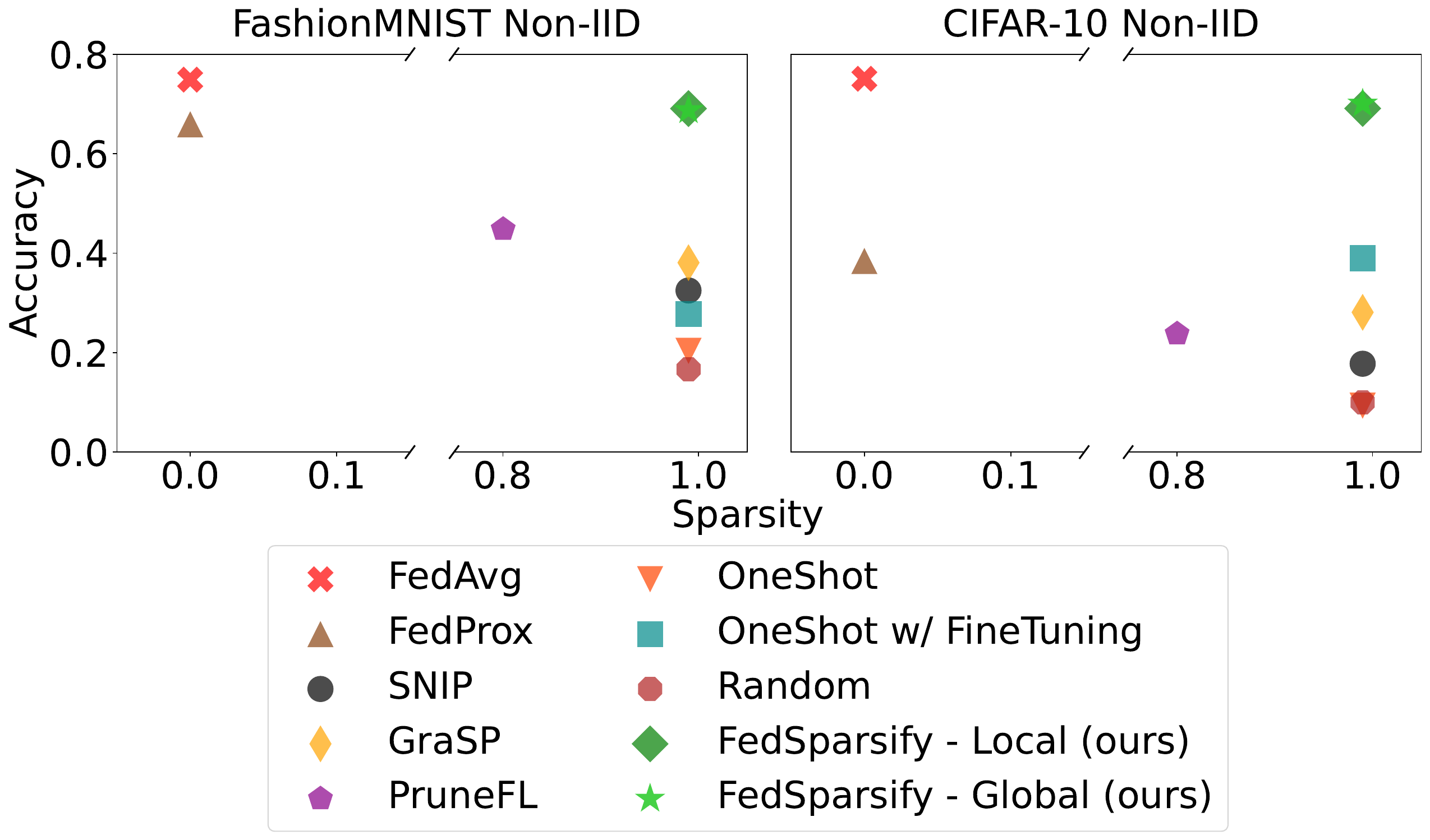}
    \caption{Test Accuracy vs.\ Sparsity for different pruning and no-pruning methods on FashionMNIST and CIFAR-10 on a federated learning environment with 10 clients and Non-IID data distribution.
    }
    \label{fig:Sparsification_IntroPlot}
\end{figure}

We systematically compare \textit{FedSparsify} to existing pruning techniques, including those that prune the model right after the initialization stage~\cite{lee2018snip,wang2019picking}, or dynamically through aggressive pruning and regrowing the model later during training~\cite{jiang2022model}, as well as non-pruning baselines~\cite{mcmahan2017communication,liu2020accelerating,li2020federatedopt}.
We also consider an adaptation of one-shot pruning in federated settings and show that federated fine-tuning is a critical step for recovering lost model performance, validating previous observations in centralized settings~\cite{han2015learning}. Although one-shot pruning is not communication efficient, we consider it to be a strong baseline for our evaluation due to its superior performance in effectively learning extremely sparse and highly accurate models. 
Compared to all these approaches, \textit{FedSparsify} is able to learn sparsified models of similar performance to non-pruning methods and outperform alternative pruning methods (see Figure~\ref{fig:Sparsification_IntroPlot}) with a 3-fold reduction in communication costs during federated training and a 4-fold improvement in inference throughput (see  Section~\ref{sec:evaluation}).

Our work provides a simple yet effective iterative pruning strategy that reaches very-high levels of sparsity while maintaining learning performance. Our approach is competitive in terms of communication, training and inference costs. We support our claims with theoretical convergence analysis and extensive experiments on challenging federated environments. Overall, we expect our work to serve as a practical baseline for future research in federated model pruning.

\section{Background and Related Work}
\label{sec:related}

Federated learning enables distributed training of machine/deep learning models without aggregating data at a single location. Typically, a federated learning environment~\cite{mcmahan2017communication} consists of a central server and $N$ participating clients that jointly train a model on their local private data. This learning environment, commonly referred to as a centralized federated topology~\cite{rieke2020future}, or star-topology~\cite{bellavista2021decentralised} is the focus of our work; though other topologies are possible~\cite{rieke2020future,bellavista2021decentralised}.
The server orchestrates the execution of the federation, each client receives the global model from the server, and trains the model for an assigned number of iterations. Depending on the number of participating clients and their availability, the server may delegate learning tasks to all clients or a subset. The ratio of clients selected for the task out of the total number of clients is called~\emph{participation ratio}~\cite{DBLP:journals/ftml/KairouzMABBBBCC21}. Upon completion of the assigned tasks, the server aggregates clients' local model parameters and computes a new global model.

\paragraph{Model Pruning for Centralized Training.}
Neural networks require significant memory and computation resources during training and inference. To this end, a large variety of model pruning techniques (i.e., techniques to remove neural network parameters that are not useful) have been proposed to improve model generalization and resource utilization~\cite{hoefler2021sparsity}. A simple and popular approach is to prune weights with smaller magnitudes at the end of the training~\cite{han2015deep} and with subsequent fine-tuning~\cite{han2015learning}. Further improvements include iteratively pruning or removing weights during training with increased pruning percentage at each iteration~\cite{han2015deep,frankle2018lottery,ZhuG18PruneOrNotPrune} and allowing resurrection of the pruned weights~\cite{lin2019dynamic,han2016dsd}. In this work, our sparsification scheme prunes weights at each federation round without resurrection, gradually increasing the number (percentage, sparsification rate) of pruned weights at each round.

Other approaches like SNIP~\cite{lee2018snip} and GraSP~\cite{wang2019picking} prune right after model initialization by exploiting gradient information. Both approaches construct a fixed sparse structure prior to the beginning of model training. Here, we focus on unstructured weight pruning, i.e., we prune at the level of single weight, and maintain a binary mask of the same size as model parameters to deactivate model weights during training and inference.
Techniques such as structured pruning~\cite{liu2018rethinking,li2016pruning} that are more friendly with current hardware can also be used.

 \paragraph{Model Pruning for Federated Training.}
\cite{DBLP:conf/ijcai/HuGG21,hu2022federated,ergun2021sparsified} have investigated gradient sparsification to enhance privacy guarantees and reduce communication overheads but not model sparsification. 
Analogously,~\cite{panda2022sparsefed} investigated gradient pruning to mitigate model poisoning attacks.
\cite{melas2022intrinsic,reisizadeh2020fedpaq,malekijoo2021fedzip} explored
gradient compression and quantization for communication cost reduction.~\cite{mitra2021linear} analyzed the convergence rate guarantees of pseudo-gradient sparsification on client and server in environments with full-client participation. For most of these works the primary focus is on faster convergence and communication cost reduction by pruning or quantizing gradients. In contrast, our aim is to learn highly sparse and highly accurate models for faster inference. Other works, PruneFL~\cite{jiang2022model} and FedDST~\cite{bibikar2022federated}, investigate dynamic model pruning. PruneFL starts with a pruned model and readjusts the sparsification mask by allowing periodic model regrow.
FedDST trains with a fixed sparsity budget throughout training and follows a dynamic pruning schedule, allowing regrowing of pruned parameters.
Our \textit{FedSparsify} strategy follows iterative cycles of pruning and fine-tuning, with gradually increasing sparsity.

\section{FedSparsify: Federated Purge-Merge-Tune}
\label{sec:FedSparsify}
\begin{algorithm}[htpb]
 \caption{\texttt{FedSparsify.} Global model $w$ and global mask $m$ are computed from $N$ participating clients (indexed by $k$) at round $t$ out of $T$ rounds. $E$ is the local training epochs; $s_t$ is the sparsification percentage of model weights; $purging\_mask$ is the pruning operation returning the  binary sparsification mask; the $\odot$ operator is the Hadamard product; $\mathcal{B}$ is the total number of batches per epoch; $\eta$ is the learning rate; $g_k^{(i)}$ denotes gradient of $k$\textsuperscript{th} client's objective with parameters $w_k^{(i)}$. If no sparsification is used \textit{FedSparsify-Global} is equivalent to FedAvg.}
    \label{alg:fedpurgemetune}
    \DontPrintSemicolon
    \SetKwProg{Fn}{Procedure}{:}{end}
    \Fn{Server($w^{(1)}, m^{(1)}$)}{
        \For{$t = 1$ \KwTo $T$}{
            \uIf{\textbf{FedSparsify-Global}}{
                \For{$k = 1$ \KwTo $N$}{
                    $w_k^{(t)} = Client(w^{(t)}, m^{(t)}, E, null)$
                }
                $w^{(t+1)} =  \sum_{k=1}^N\frac {|\mathcal D_k|} {|\mathcal D|} w_k^{(t)}$\\
                $m^{(t+1)} = purging\_mask(w^{(t+1)}, s_t)$\\
                ${w}^{(t+1)} = {w}^{(t+1)}\odot m^{(t+1)}$
            }

            \uIf{\textbf{FedSparsify-Local}}{
                \For{$k = 1$ \KwTo $N$}{
                    $w_k^{(t)}, m_k^{(t)} = Client(w^{(t)}, m^{(t)}, E, s_t)$
                }
                $(w^{(t+1)}, m^{(t+1)}) \coloneqq$ Compute using Eq.~\ref{eq:majority_voting}
            }
        }
        \textbf{return} $w^{(t+1)}$
    }
    \;
    \Fn{Client($w, m, E, s_t$)}{
    $w_k^{(0)} = w$\\
    $S = E*\mathcal B$\\
    \For{$i = 0 \ \KwTo\ S$}{
            $w_k^{(i+1)} = w_k^{(i)} -\eta g_k^{(i)}\odot m$
        }
        \uIf{\textbf{FedSparsify-Local}}{
            $m_{k} = purging\_mask \left(w_k^{(S)}, s_t\right)$ \\
            \textbf{return} $\left(w_k^{(S)}, m_k\right)$
        }
        \textbf{return} $w_k^{(S)}$
    }

\end{algorithm}

\textit{FedSparsify} follows an iterative pruning schedule that performs model pruning based on weight magnitude at different learning tiers, clients or server. The entire procedure is summarized in Algorithm~\ref{alg:fedpurgemetune}.

\paragraph{Weight Magnitude-based Pruning.} %
Neural networks often have millions of parameters, but not all parameters influence the outcome/predictions equally. A simple and surprisingly effective proxy to identify weights with small effect on the final outcome is based on the weights' magnitude~\cite{han2015deep,frankle2018lottery}. Weights with magnitudes lower than some threshold can be removed or set to zero without penalizing performance. We choose this threshold  based on the number of parameters to be pruned (or prune percent, $s_t$).
We prune parameters whose weight magnitude is in the bottom-$s_t$\% in an unstructured way, i.e., considering the magnitude of each parameter separately. Our approach is modular and  other model pruning approaches that prune groups of parameters, i.e., structured pruning~\cite{liu2018rethinking}, based on magnitude can also be readily used.

\paragraph{Pruning Schedule.}
A critical step in our approach is how often and how many parameters to prune during federated training. Pruning too many parameters earlier in training can cause irrecoverable damage to the performance~\cite{frankle2018lottery}, and pruning too late leads to increased communication costs. To balance this, we prune iteratively, by gradually reducing the number of trainable parameters. Fine-tuning after pruning often improves the performance and allows pruning of more parameters while preserving performance~\cite{han2015learning,ZhuG18PruneOrNotPrune,frankle2018lottery}. Therefore, model pruning at the end of each federation round is a natural choice since clients can fine-tune the aggregated pruned global model during the next federation round. Intuitively, two pruning strategies are possible --- prune locally at the clients before aggregation (FedSparsify-Local), or globally at the server after aggregation (FedSparsify-Global). We explore both strategies. Once a parameter is pruned, it never rejoins training (i.e., no network/weight regrowth is allowed). Motivated by~\cite{ZhuG18PruneOrNotPrune}, we adapt the iterative exponentially weight pruning formula of a standalone model in a centralized setting into a federated environment to the following federated pruning schedule:
\begin{equation}
    s_t = S_T + \left(S_0 - S_T\right)  \left(1 - \frac{F\lfloor t/F\rfloor-t_0}{T-t_0}\right)^ n
    \label{eq:pruning_schedule}
\end{equation}
where $t$ is the federation round, $s_t$ is the sparsification percentage at rount $t$, $S_T$ is the final desired sparsification, $S_0$ is the initial sparsification percentage, $t_0$ is the round at which pruning starts, $T$ is the total number of rounds, and $F$ is the pruning frequency (e.g., $F=1$ prunes at every round, while $F=5$ prunes every 5 rounds). The exponent $n$ controls the rate of sparsification. A higher $n$ leads to aggressive sparsification at the start of training, and a smaller $n$ to more sparsification towards the end of training; we use $n=3$ in our experiments. Overall, this formula provides an interplay between communication cost, model sparsity and learning performance (see also Appendix~\ref{sec:Appendix_FederatedExperimentsSpecifications}).

\paragraph{FedSparsify-Local.} Model pruning takes place at each client after local training is complete. Each client sends its model, $w_k$, to the server along with the associated binary sparsification  masks, $m_k$. The server may aggregate the local models using FedAvg. However, as the number of clients increases, it is increasingly unlikely that a particular weight will be zero for all clients. This results in slow sparsification rates. To address this, we aggregate local models using our proposed Majority Voting scheme, where a global model parameter is zeroed out only if less than half of local models' masks preserve it. Otherwise, the standard weighted average aggregation rule applies. Formally:
\begin{equation}
    \footnotesize 
    \begin{aligned}
    [m]_i = \begin{cases}
                1 & \sum_{k}^{N} [m_{k}]_{i} \geq \frac{N}{2}\\
                0 &\text{otherwise}
    \end{cases};
    \enspace
    w = m \odot \left(\sum_{k}^{N} \frac{|\mathcal{D}_k|}{\mathcal{|D|}} w_{k} \right)
    \end{aligned}
    \label{eq:majority_voting}
\normalsize
\end{equation}

where $[\cdot]_i$ is the parameter value at the $i$\textsuperscript{th} position, $w$ is the global model, $N$ is the number of clients participating in the current round, and $m_k$ is the local binary mask of client $k$.

\paragraph{FedSparsify-Global.} Model pruning occurs at the server right after participating clients' models are aggregated and the sparse structure is maintained throughout local training.

\paragraph{}
FedSparsify-Global and FedSparsify-Local pruning differ primarily on the pruning tier. In FedSparsify-Global, the server prunes the global model after aggregation.
This is in contrast with FedSparsify-Local, where the clients prune their local model after local training is complete and share their local binary masks with the server, and the server aggregates the models using the Majority Voting scheme. In both schemes, there is no mask disagreement during local training, namely all clients update the same set of model parameters due to the shared global mask.
Intuitively, FedSparsify-Local prunes using only the local client information, whereas FedSparsify-Global prunes after aggregation and hence uses information from all clients. Due to this, we expect FedSparsify-Global to be slightly better than FedSparsify-Local in terms of performance. This is validated in our empirical evaluation (Section~\ref{sec:evaluation}). We observe that FedSparsify-Global outperforms FedSparsify-Local across almost all federated environments, especially in the more challenging Non-IID environments. FedSparsify-Local may be useful in asynchronous settings but our experiments concern only synchronous settings.

\section{FedSparsify Convergence Analysis}\label{sec:convergence}

Borrowing the notation from Section~\ref{sec:FedSparsify} and Appendix~\ref{appendix:proof}, we analyze the convergence rate for \textit{FedSparsify} when $|\mathcal D_k|=|\mathcal D|/N, \forall\ k$, i.e., equal weights for each client and participation ratio is 1. These relaxations are made to simplify the analysis, but are not critical to the proof. See~\cite{li2019convergence,hu2022federated} regarding the treatment of partial participation at each round, and~\cite{li2019convergence,jiang2022model} for analysis with a weighted average. We make the same assumptions as~\cite{hu2022federated,jiang2022model}, which are stated in Appendix~\ref{appendix:proof}. 
\begin{theorem}
\label{thm:convergence}
    If assumptions \ref{assn:1}-\ref{assn:7} hold and the learning rate, $\eta< (4\sqrt 2LS)^{-1}$, then
    the parameters obtained at the end of each federation round of \textit{FedSparsify} algorithm satisfy
    \begin{align*}
        \frac 1 T\sum_{t=1}^T
        & \left\| 
            m^{(t)}   \odot {\nabla f\left(w^{(t)}\right)} 
        \right\|^2
        \leq
        2\eta L \left[\left(1+4L\eta S\right)\sigma^2
            +\left(16L\eta S^2\right)\epsilon^2\right]
        \\
        & 
        +\frac{4}{T\eta S} \E \left[f\left(w^{(1)}\right) - f\left(w^{(*)}\right)\right]
        + \frac{4}{T \eta S} \sum_{t=1}^T L_p
                \left\|w^{(t+1)} -w^{(t+1)}\odot m^{(t)}\right\|
        \nonumber
    \end{align*}
    \noindent where $L, L_p, \sigma, \epsilon$ are defined in assumptions, S and T are number of local updates and federations rounds resp., $w^{(t+1)}\odot m^{(t)}\coloneqq \frac 1 N \sum_{k=1}^N w_k^{(t,S)}$, i.e., parameters right before sparsification is done and $w^{(*)}$ is the optimal parameter of sparsity $S_T$.
\end{theorem}
The proof is provided %
in Appendix~\ref{appendix:proof}. By setting the learning rate, $\eta= \mathcal O(\frac {1} {\sqrt{TS}} )$, we can see that the sequence converges at
$\mathcal O(\frac {1} {\sqrt{TS}} )$ rate, which is the same as FedAvg under the same assumptions~\cite{li2019convergence,yu2019linear}.
However, compared to training with FedAvg, the bound for \textit{FedSparsify} has an additional term---the magnitude of the difference of weights before and after pruning. This may explain the effectiveness of pruning weights with the lowest magnitude, as pruning these would minimize the additional term, i.e., the magnitude of the difference of weights, compared to other choices. 

 We can further upper bound the difference by observing that $m^{(t)}$ describes the non-zero parameters in t\textsuperscript{th} federation round, then 
\[
    \left\|w^{(t+1)}- w^{(t+1)}\odot  m^{(t)}\right\|
    \leq \left\|w^{(t+1)} \odot m^{(t)}\right\| 
\]
and assuming the magnitude of neural network parameters is upper bounded by B (as assumed in~\cite{jiang2022model}). However, this naive upper bound ignores that we prune parameters with the lowest magnitude in \textit{FedSparsify-Global}, which can lead to a tighter bound for \textit{FedSparsify-Global}. Note that $w^{(t+1)} - w^{(t+1)}\odot m^{(t)}$ will be $0$ everywhere except for the indexes marked for pruning, i.e., the smallest entries before $t+1$\textsuperscript{th} round. Note that exactly $\lfloor|w|\times s_{t+1}\rfloor-\lfloor|w|\times s_{t}\rfloor$ will be non-zero, giving a tighter bound. 
\begin{align*}
    \left\|w^{(t+1)}- w^{(t+1)}\odot  m^{(t)}\right\|
    &\lesssim \left\|w^{(t+1)}\odot  m^{(t)}\right\|(s_{t+1}-s_t)
    \\
    &\lesssim \left\|w^{(t+1)}\right\|\frac{s_{t+1}-s_t}{1-(s_{t+1}-s_t)}    
\end{align*}
We use $\lesssim$ instead of $\leq$ to be explicit that integer effects are ignored. 
In the case of FedSparsify-Local and majority voting, we remove parameters based on whether most clients agree. Thus, the pruned parameter values are not necessarily the smallest, and the above-discussed bound may not hold.

\section{Evaluation}
\label{sec:evaluation}
We compare \textit{FedSparsify} against a suite of pruning algorithms performing model sparsification at different stages of federated training, as well as non-pruning methods\footnote{\url{https://github.com/dstripelis/FedSparsify}}.

\paragraph{Baselines.} 
We compare our \textit{FedSparsify-Global} and \textit{FedSparsify-Local} approaches against pruning at initialization schemes that sparsify the global model prior to federated training, such as \textit{SNIP}~\cite{lee2018snip} and \textit{GraSP}~\cite{wang2019picking}), and a dynamic pruning scheme that prunes and resurrects model parameters during training, \textit{PruneFL}~\cite{jiang2022model}. We also conceptualize a \textit{OneShot pruning} with fine-tuning baseline for federated training based on~\cite{han2015learning}. Lastly, we validate the benefits of magnitude-based pruning by substituting it with random weight pruning in FedSparsify-Global; we refer to this scheme as~\textit{Random}. 

SNIP~\cite{lee2018snip} and  GraSP~\cite{wang2019picking} construct a fixed sparse model prior to the beginning of federated training. Following previous works~\cite{jiang2022model,bibikar2022federated}, we apply the schemes in a federated setting by randomly picking a client at the beginning of training to create the initial sparsification mask, and enforce it globally throughout training.

PruneFL~\cite{jiang2022model} aims to maximize the reduction of empirical risk per training time. It prunes before training as well as during training. During training, PruneFL identifies parameters to prune based on the ratio of gradients magnitude and execution time. We follow the training and pruning configurations suggested in the original work.
At the start of training, PruneFL picks a client at random from the federation to learn the initial pruning mask after 5 reconfigurations. We perform global mask readjustment every 50 rounds and sparsification ratio at round $t$ is set to $s \times 0.5^\frac{t}{1000}$ with $s=\{0.3, 0.8\}$; $0.3$ is the recommended value.

We adapt the one-shot pruning~\cite{han2015learning} approach originally proposed for centralized settings into federated settings by training the original dense model for a specific number of rounds and prune the global model only once (at the server) to the desired sparsity. Similar to other pruning approaches, to restrict training on the non-pruned weights across all clients, the server shares the global model's sparsification mask with every client to mask their local updates. For our evaluation, we consider two variations of this approach, one where the global model is not fine-tuned after pruning (\textit{OneShot}) and one where it is fine-tuned after pruning (\textit{OneShot w/ FineTuning}) for the last 10 rounds.

Finally, we consider FedAvg with Vanilla SGD~\cite{mcmahan2017communication}, FedAvg with Momentum SGD~\cite{liu2020accelerating}, referred to as FedAvg (MFL), and FedProx \cite{li2020federatedopt} as the non-pruning baselines. For FashionMNIST we use FedAvg with SGD and FedProx, and FedAvg (MFL) and FedProx for CIFAR-10 and CIFAR-100.

\paragraph{Federated Models \& Environments.}
We use FashionMNIST, CIFAR-10 and CIFAR-100 as the benchmark datasets. We train a 2-layer fully-connected (FC) network for FashionMNIST, a 6-layer convolutional network (CNN) for CIFAR-10 and a VGG-16 network for CIFAR-100, with 118,282, 1,609,930 and 14,782,884 trainable parameters, respectively. We create four federated environments for each dataset by varying the number of clients (10 and 100 clients) and data distribution (IID and Non-IID) at the clients.
We generate the IID data distributions by randomly partitioning the dataset into 10 and 100 chunks~\cite{mcmahan2017communication}. We create non-IID data distributions by skewing the label distribution~\cite{kairouz2019advances}. A  subset of classes are assigned to each client similar to \cite{zhao2018federated,hsu2019measuring}: 2 classes (out of 10) for FashionMNIST, 5 classes (out of 10) for CIFAR-10, and 50 classes (out of 100) for CIFAR-100. All clients participate in every round in the environments of 10 clients. In the 100 clients environments, 10 clients are randomly selected at each round (i.e., 0.1 participation rate). All training hyperparameters and federated environments are presented in Appendix~\ref{sec:Appendix_FederatedExperimentsSpecifications}.

\begin{figure}[htbp]
  \centering
  \subfloat[CIFAR-100 (VGG)]{
  \includegraphics[width=.95\linewidth, trim={0, 7.6cm, 0, 0}, clip]{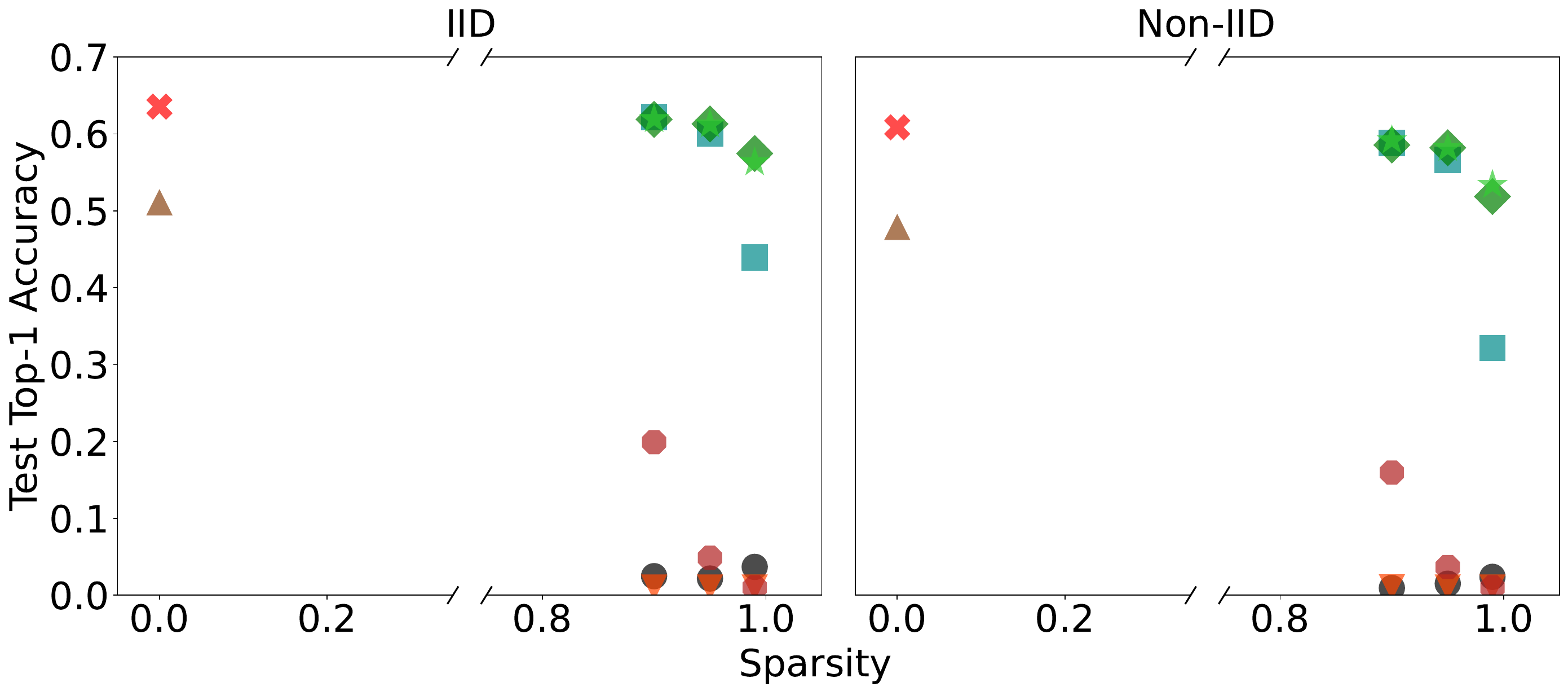}
  \label{subfig:MainPaper_CIFAR100_SparsificationSchemesSparsityComparison_10clients}
  }
  \\[-0.5ex]
  \subfloat[CIFAR-10 (CNN)]{
  \includegraphics[width=.95\linewidth, trim={0, 8.95cm, 0, 0}, clip]{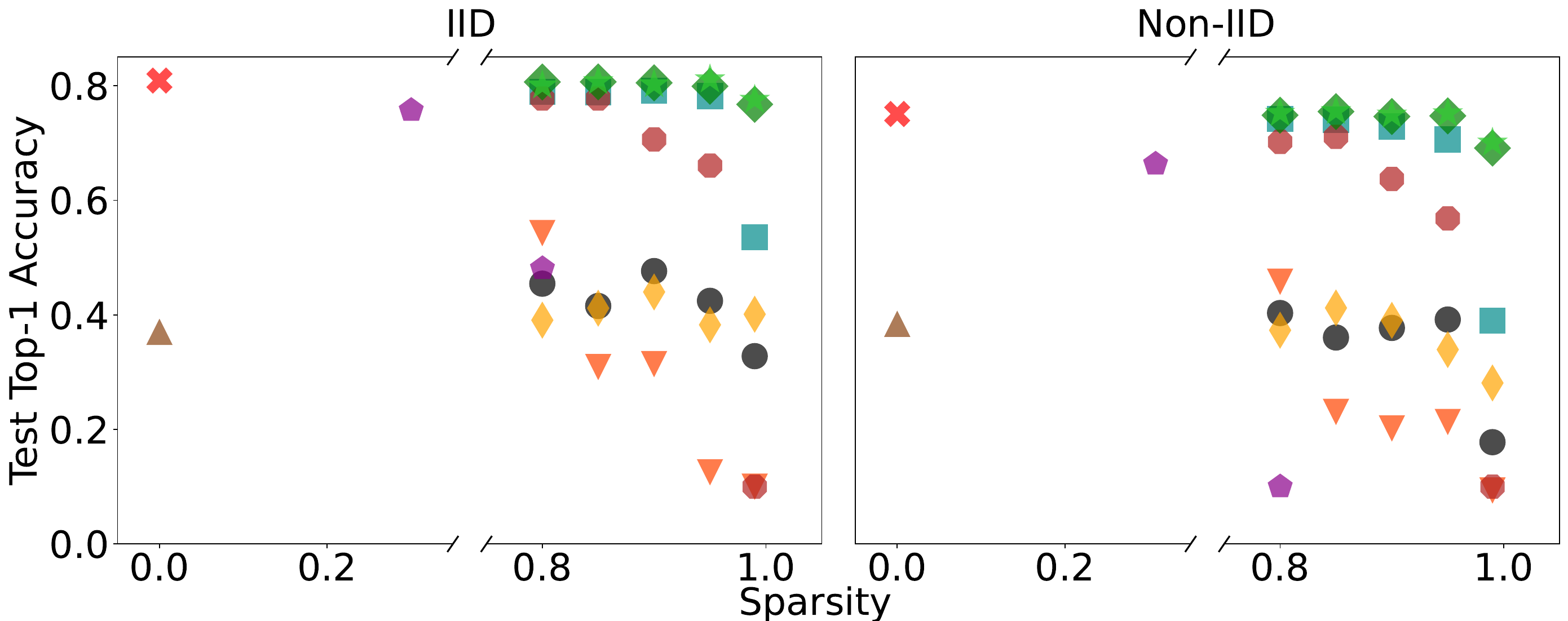}
  \label{subfig:MainPaper_CIFAR10_SparsificationSchemesSparsityComparison_10clients}
  }
  \\[-0.5ex]
  \subfloat[FashionMNIST (FC)]{
    \includegraphics[width=.95\linewidth]{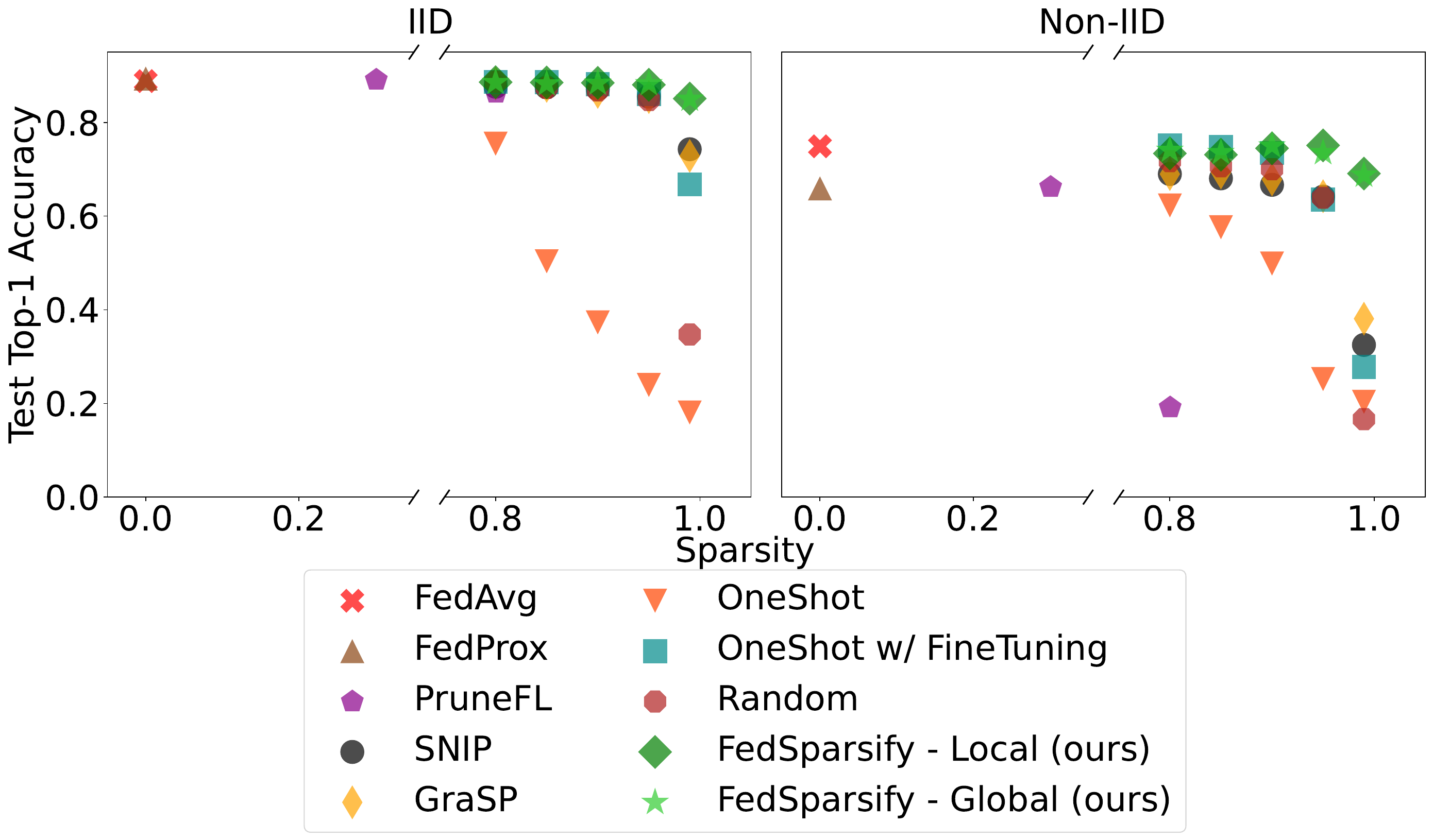}
  \label{subfig:MainPaper_FashionMNIST_SparsificationSchemesSparsityComparison_10clients}
  }
  
  \caption{Sparsity vs. Test Accuracy for 10 clients. FedSparsify outperforms pruning alternatives, and is comparable to no-pruning.} %
  \label{fig:MainPaper_SparsificationSchemesComparison_10Clients}
\end{figure}

\begin{figure}[htpb]
  \centering
  \subfloat[CIFAR-100 (VGG)]{
  \includegraphics[width=.95\linewidth, trim={0, 7.6cm, 0, 0}, clip]{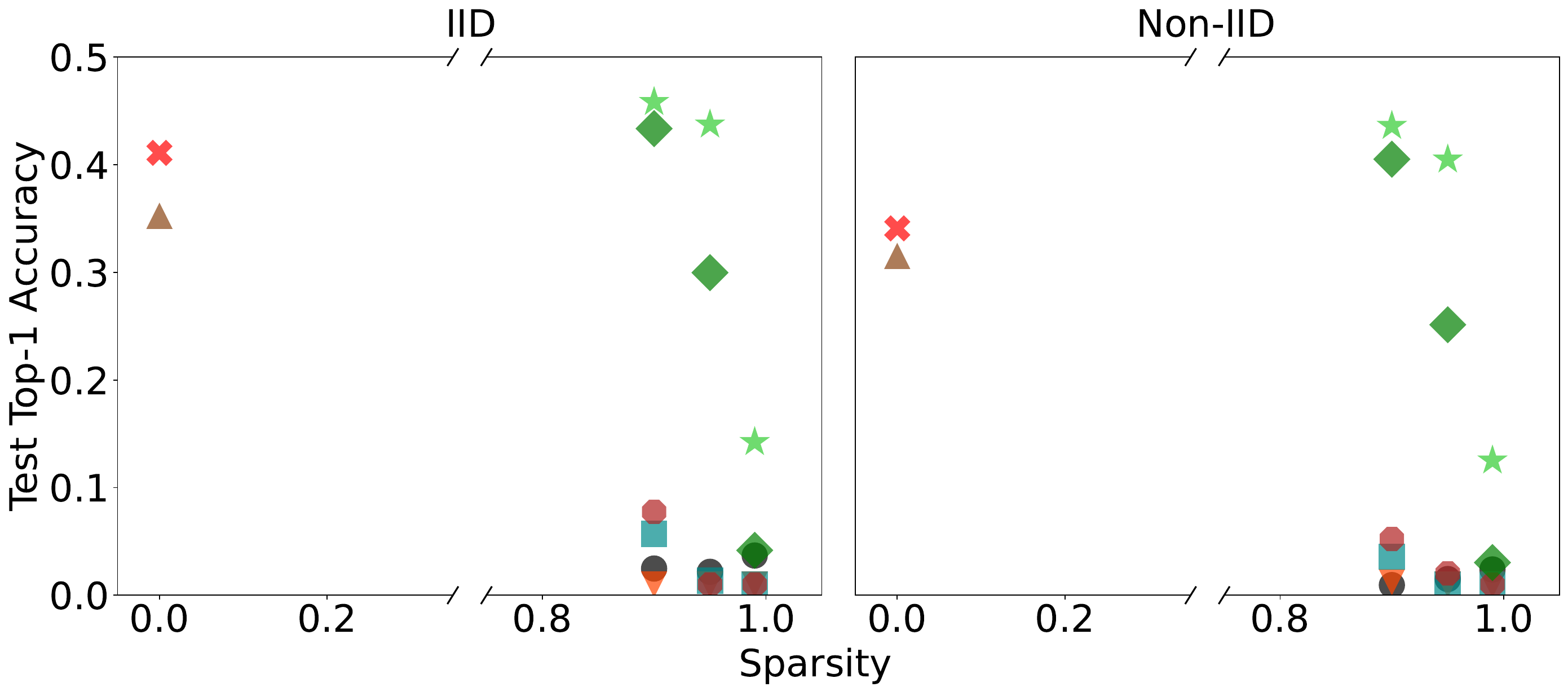}
  \label{subfig:MainPaper_CIFAR100_SparsificationSchemesSparsityComparison_100clients}
  }
  \\[-0.5ex]
  \subfloat [CIFAR-10 (CNN)]{
  \includegraphics[width=.95\linewidth, trim={0, 8.95cm, 0, 0}, clip]{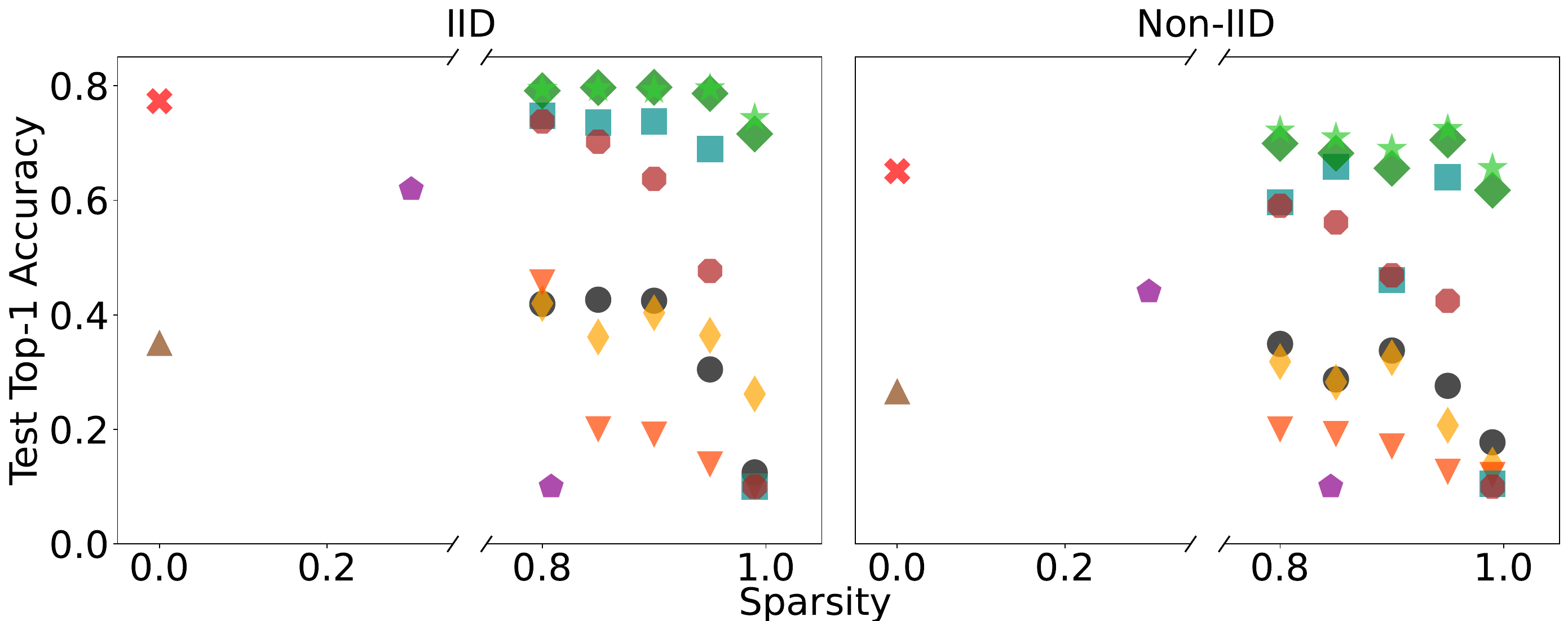}
  \label{subfig:MainPaper_CIFAR10_SparsificationSchemesSparsityComparison_100clients}
  }
  \\[-0.5ex] 
  \subfloat [FashionMNIST (FC)]{
  \includegraphics[width=.95\linewidth, trim={0, 0, 0, 0}, clip]{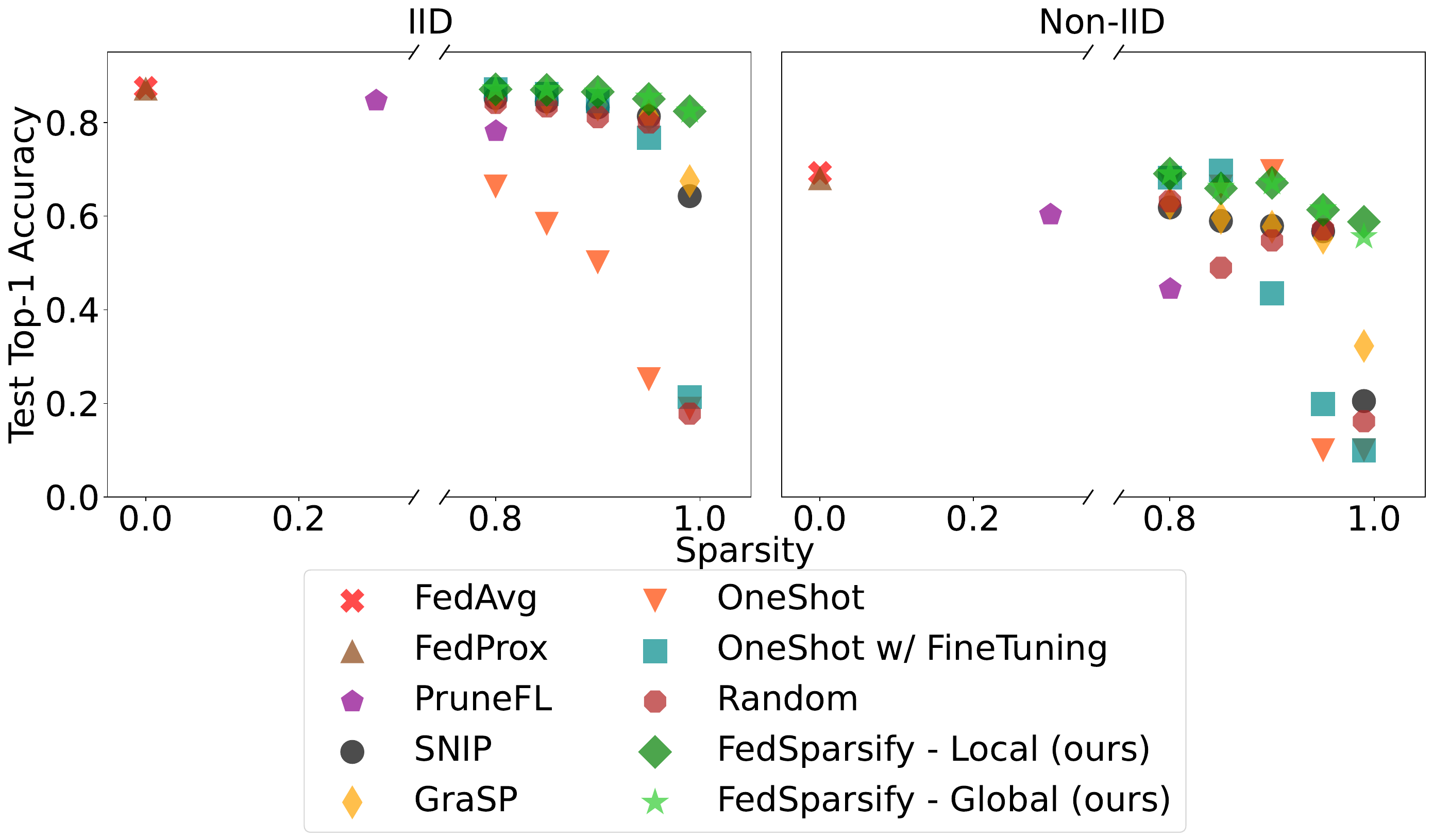}
  \label{subfig:MainPaper_FashionMNIST_SparsificationSchemesSparsityComparison_100clients}
  }
   \caption{Sparsity vs. Test Accuracy for 100 clients (0.1 participation rate). FedSparsify outperforms pruning alternatives and is comparable or better to no-pruning (particularly in non-IID domains).}
  \label{fig:MainPaper_SparsificationSchemesComparison_100Clients}
\end{figure}

\begin{figure}[htpb]
  \centering
  \subfloat[10 Clients, Participation Ratio: 1]{
  \includegraphics[width=\linewidth, trim={0, 7.4cm, 0, 0}, clip]{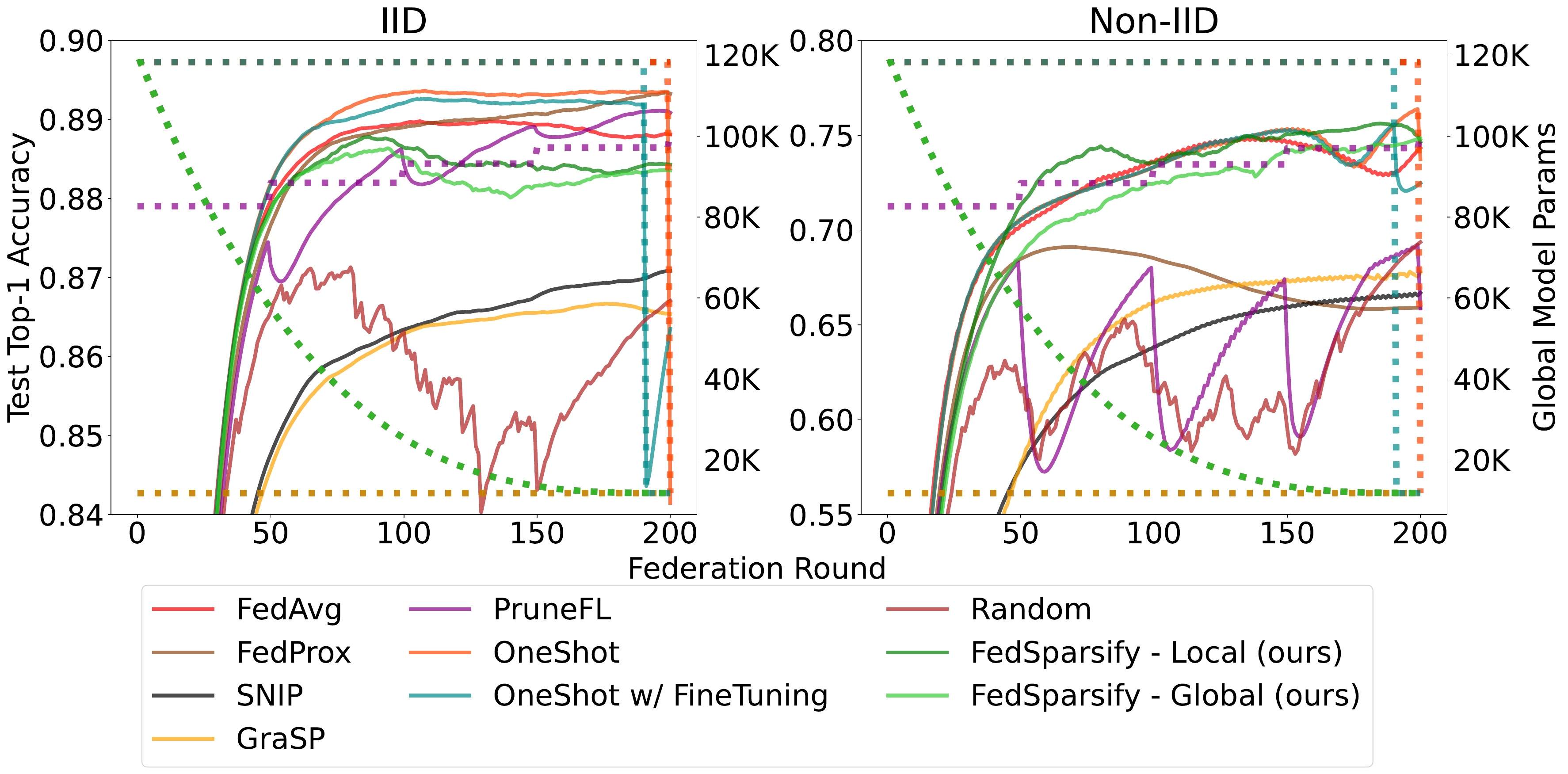}
  \label{subfig:FashionMNIST_MainBody_Convergence_FederationRoundsComparison_10clients}
  }
  \\[-0.5ex]
  \subfloat[100 Clients, Participation Ratio: 0.1]{
  \includegraphics[width=\linewidth]{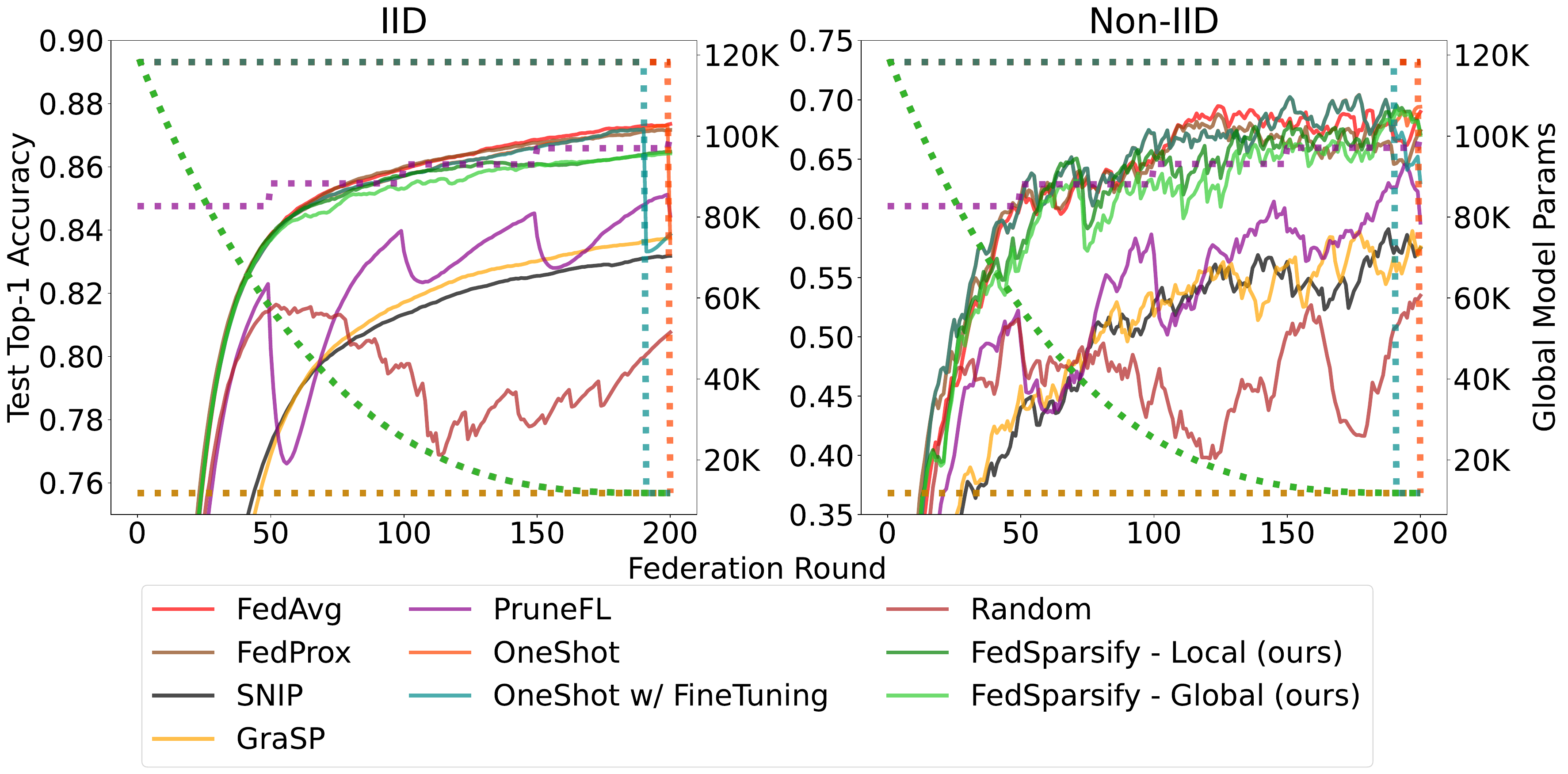}
  \label{subfig:FashionMNIST_MainBody_Convergence_FederationRoundsComparison_100clients}
  }
   \caption{FashionMNIST global model convergence in terms of Federation Rounds vs. Accuracy (left y-axis, solid lines) and Parameters progression (right y-axis, dashed lines) for 10 and 100 clients. Sparsity for PruneFL is set to 0.3 and for all other methods to 0.9.}
  \label{fig:FashionMNIST_MainBody_Convergence_FederationRoundsComparison}
\end{figure}

\paragraph{Evaluation Criteria.}
Our goal is to develop federated training strategies that can learn a global model with the highest achievable accuracy at high sparsification rates. To this end, we evaluate the trade-off between model sparsity and learning performance (i.e., accuracy). 
Figures~\ref{fig:MainPaper_SparsificationSchemesComparison_10Clients} and~\ref{fig:MainPaper_SparsificationSchemesComparison_100Clients} show the accuracy on a held-out test set at different sparsities (0.8, 0.85, 0.9, 0.95, 0.99) for FashionMNIST and CIFAR-10, and (0.9, 0.95, 0.99)
for CIFAR-100.
We also evaluate model operations and model throughput in Table~\ref{tbl:MainPaper_CIFAR_ModelsComparison}. We report model convergence with respect to (w.r.t.) federation rounds and global model size reduction for FashionMNIST in Figure~\ref{fig:FashionMNIST_MainBody_Convergence_FederationRoundsComparison}. Convergence plots for all other environments and communication cost of federated training are presented in Appendix~\ref{sec:Appendix_FederatedExperimentsSpecifications}. We do not measure convergence w.r.t. computation or wall-clock time speed-up, since we do not employ any dedicated hardware accelerators to perform sparse operations.

\paragraph{FashionMNIST Results.} Figures~\ref{subfig:MainPaper_FashionMNIST_SparsificationSchemesSparsityComparison_10clients} and~\ref{subfig:MainPaper_FashionMNIST_SparsificationSchemesSparsityComparison_100clients} show the performance of different methods at different sparsity for the FashionMNIST environments with 10 and 100 clients, respectively, on IID and Non-IID data distributions. The more complex the learning environment is (cf.\ IID vs Non-IID), the lower the final accuracy of the global model is for both pruning and non-pruning schemes. All sparsification methods have similar performance at moderate sparsification (i.e., 0.8, 0.85, 0.9) and IID distribution. However, for extreme sparsities  (i.e., 0.95, 0.99) and more challenging data distributions (Non-IID), other sparsification methods underperform and, in some cases, cannot even learn a model of reasonable performance (e.g., SNIP, GraSP and One-Shot).  %

Even though SNIP and GraSP are capable of training with reduced communication costs compared to other approaches (cf. Figure~\ref{fig:Appendix_Convergence_TransmissionCost} in Appendix~\ref{sec:Appendix_Additional_Evaluation}) due to restricting model training to a predetermined sparsified network from the very beginning they incur a substantial performance drop compared to our progressive sparsification schemes. We attribute this performance degradation to the binary mask learned over the local dataset of a randomly selected client, which may not follow the global data distribution, especially in the case of Non-IID environments; similar observations were also reported in~\cite{bibikar2022federated}. Similarly, the degraded learning performance of PruneFL in the Non-IID settings is due to the random client selection at the start of federated training for constructing the initial sparsification mask. 
A noteworthy outcome of our experiments is the effect of fine-tuning in the case of one-shot pruning. Fine-tuning improves the final performance  by a large margin in IID (0.2 vs 0.7 at 0.99 sparsity) and Non-IID settings (0.25 vs 0.67 at 0.95 sparsity).

Figures~\ref{subfig:FashionMNIST_MainBody_Convergence_FederationRoundsComparison_10clients} and~\ref{subfig:FashionMNIST_MainBody_Convergence_FederationRoundsComparison_100clients} show global model convergence and its total number of parameters during training. FedAvg, FedProx, OneShot and OneShot w/ FineTuning have a constant model size (overlapping top dashed lines), except for the latter two approaches for which pruning occurs at round 190 and 200, respectively, and hence the sudden size drop. Similarly, SNIP and GraSP train on a pruned model of constant size (overlapping bottom dashed lines), since the initial training model is already sparsified. All progressive sparsification schemes (FedSparsify, Random) have a logarithmically decreasing model size (mid-low overlapping decreasing dashed lines) while the dynamic pruning scheme (PruneFL) has a step-like increasing model size.

PruneFL's performance drops every 50 federation rounds when the model expands parameters. The effect is stronger in the Non-IID environment. As expected, OneShot pruning behaves similarly to the non-pruning baselines until pruning. The accuracy drops substantially when the model is pruned at round 200 for OneShot and 190 for OneShot w/ FineTuning due to the removal of a large number of parameters that affects model outcomes. Interestingly, OneShot w/ FineTuning recovers some of the lost performance during federated fine-tuning in the last 10 rounds. Finally, Random pruning does not suffer during early federated training or at lower sparsities but fails to perform towards the end of training, indicating the usefulness of magnitude-based pruning.

\begin{table*}[htbp]
    \centering
    \footnotesize
    \subfloat[CIFAR-10\label{tbl:CIFAR10_ModelsComparison}]{
    \centering
    \setlength\tabcolsep{1.5pt}
    \begin{tabular}{@{}cccccccc@{}}
        \toprule
        Sparsity & Accuracy & Params & Model Size (MBs) & C.C. (MM) & Inf.Latency & Inf.Iterations & Inf.Throughput \\ \midrule
        0.0 & 0.75 & 1,609,930 & 5.903 & 6,441 & 115 & 4,145 & 8,831 \\
        0.8 & 0.752 & 322,370 & 1.54 (x3.83) & 2,596 (x2.48) & 61 (x1.88) & 7,812 (x1.88) & 16,651 (x1.88) \\
        0.85 & 0.755 & 241,874 & 1.178 (x5.01) & 2,356 (x2.73) & 51 (x2.25) & 9,222 (x2.22) & 19,660 (x2.22) \\
        0.90 & 0.749 & 161,377 & 0.802 (x7.36) & 2,116 (x2.97) & 43 (x2.67) & 10,975 (x2.64) & 23,399 (x2.64) \\
        0.95 & 0.751 & 80,881 & 0.415 (x14.22) & 1,875 (x3.43) & 32 (x3.59) & 14,682 (x3.54) & 31,306 (x3.54) \\
        0.99 & 0.7 & 16,484 & 0.104 (x56.75) & 1,683 (x3.82) & 27 (x4.25) & 17,707 (x4.27) & 37,763 (x4.27) \\
        \bottomrule
    \end{tabular}
    }

    \subfloat[CIFAR-100\label{tbl:CIFAR100_ModelsComparison}]{
    \centering
    \setlength\tabcolsep{1.5pt}
    \begin{tabular}{@{}cccccccc@{}}
        
        \toprule
        Sparsity & Accuracy & Params & Model Size (MBs) & C.C. (MM) & Inf.Latency & Inf.Iterations & Inf.Throughput \\ \midrule
        0.0 & 0.6091 & 14,782,884 & 54.581 & 946,104 & 524 & 919 & 1945 \\ %
        0.90 & 0.5886 & 1,485,892 & 7.556 (x7.22) & 314,253 (x3.01) & 171 (x3.06) & 2801 (3.04) & 5962 (x3.06) \\ %
        0.95 & 0.5817 & 747,170 & 3.978 (x13.72) & 279,150 (x3.38) & 134 (x3.91) & 3575 (x3.89) & 7613 (x3.91) \\ %
        0.99 & 0.551 & 156,193 & 0.881 (x61.95) & 251,067 (x3.76) & 65 (x8.06) & 7294 (x7.93) & 15549 (x7.99) \\ %
        \bottomrule
    \end{tabular}
    }
    
    \caption{Comparison of sparse (\textit{FedSparsify-Global}), and non-sparse (\textit{FedAvg}) federated models in the CIFAR-10 and CIFAR-100 Non-IID environments with 10 clients. Inference evaluations are done on models obtained at the end of training. Sparsity 0.0 represents FedAvg.
    C.C.\ is the communication cost in millions (MM) of parameters exchanged. Inference efficiency is measured by the mean processing time per batch (Inf.Latency - ms/batch), the number of iterations (Inf.Iterations), and processed examples per second (Inf.Throughput - examples/sec). Values in parenthesis show the reduction factor (model size, communication cost and inference latency) and increase/speedup factor (inference iterations and throughput) compared to non-pruning.}

    \label{tbl:MainPaper_CIFAR_ModelsComparison}
\end{table*}

\paragraph{CIFAR-10 Results.} 
As shown in Figures~\ref{subfig:MainPaper_CIFAR10_SparsificationSchemesSparsityComparison_10clients} and~\ref{subfig:MainPaper_CIFAR10_SparsificationSchemesSparsityComparison_100clients} FedSparsify outperforms existing federated pruning approaches, while being able to learn sparse models at extreme sparsification rates (e.g., 0.9-0.99) and often the performance is similar or better than the non-pruning FedAvg (MFL) baselines (e.g., Non-IID environment in Figure~\ref{subfig:MainPaper_CIFAR10_SparsificationSchemesSparsityComparison_10clients}).
Similar to the  FashionMNIST results, we attribute the performance drop of pruning at initialization schemes to their need to remove a large proportion of the network's trainable weights at the beginning of training based on the local dataset of a randomly chosen client that may not be representative of the global dataset.
Similarly, PruneFL also relies on an initial randomly selected sparsification mask, and cannot recover performance even after model regrow during federated training.

Interestingly, the Random pruning scheme is a strong baseline with comparable and often better performance compared to existing pruning methods but worse to one-shot pruning approaches. However, at extreme levels of sparsity, random pruning does not learn as the remaining model weights are crucial and pruning randomly may have an irreversible, negative effect on the final model performance.
OneShot pruning with finetuning has comparable performance as our FedSparsify schemes generally but fails to learn a model of reasonable performance  at very extreme levels of sparsity, i.e., 99\%. Even though fine-tuning can help recover the lost performance at almost all sparsification degrees it fails to do so at such extreme levels. However, the iterative pruning removes parameters gradually and hence can maintain performance even at extreme sparsity levels. 

The results on 100 clients are similar. FedSparsify outperforms alternative pruning methods with performance comparable or better to non-pruning. In the more challenging Non-IID environment, FedSparsify-Global performs slightly better than FedSparsify-Local (see Figure~\ref{subfig:MainPaper_CIFAR10_SparsificationSchemesSparsityComparison_100clients}).

\paragraph{CIFAR-100 Results.} Figures~\ref{subfig:MainPaper_CIFAR100_SparsificationSchemesSparsityComparison_10clients} and~\ref{subfig:MainPaper_CIFAR100_SparsificationSchemesSparsityComparison_100clients} show that both FedSparsify schemes perform comparably or better than the dense model baselines. SNIP, GraSP, and PruneFL are not able to learn a model of acceptable performance (better than random performance) within the allocated training budget of 100 federation rounds (we only plot SNIP since all three methods behave similarly). 
These methods may require a greater number of training iterations or parameters.
To validate this, we ran SNIP and GraSP in a centralized setting and observed that as the sparsification degree increases, so does the amount of training iterations needed to learn a good model. PruneFL required many more rounds and frequent communication (e.g., every 4 local update steps), as reported in the original work, whereas in our setting we synchronize models every 4 epochs.

We observe a substantial drop in accuracy from 10 clients to 100 clients ($\sim$60\% in Figure~\ref{subfig:MainPaper_CIFAR100_SparsificationSchemesSparsityComparison_10clients} vs.\ $\sim$40\% in Figure~\ref{subfig:MainPaper_CIFAR100_SparsificationSchemesSparsityComparison_100clients}).
This degradation could be due to the reduced number of training examples the global model is trained with at each round, 50000 samples for 10 clients, but only 5000 samples for 100 clients (since only 10 of the 100 clients participate). This data scarcity per training round also affects the OneShot w/ FineTuning method. Even though fine-tuning helps recover some of the lost model performance in the case of 10 clients, its effect is minor on the 100 clients. However, FedSparsify outperforms even FedAvg and FedProx in the 100 clients' environments.
We attribute this behavior to the regularization effect of gradual pruning during training~\cite{hoefler2021sparsity}. Finally, at high sparsity degrees, FedSparsify-Global performs better than FedSparsify-Local.
We posit FedSparsify-Global is better suited to efficiently learn extremely sparse models from large networks.

\paragraph{Sparsification Efficiency.} The main goal of our sparsification scheme is to improve federated models' inference efficiency while being equally performant as non-pruning methods.
Table~\ref{tbl:MainPaper_CIFAR_ModelsComparison} shows a quantitative comparison of this claim. The Table reports the performance of non-pruning (FedAvg) and sparsified models learned using our FedSparsify-Global approach in the non-IID environments with 10 clients for CIFAR-10 and CIFAR-100.
Following previous work on benchmarking inference efficiency of sparsified models~\cite{pmlr-v119-kurtz20a,DBLP:journals/corr/abs-2111-13445}, we record the total number of batches (iterations) completed by the sparsified model within an allocated execution time and compute the number of items processed per second (throughput - items/sec) and the processing time per batch (ms/batch).
Compared to dense CNN and VGG models, the learned sparse models have significant efficiency improvements. Sparse models at 0.99 sparsity provide a 4-fold (CNN) and 8-fold (VGG) improvement w.r.t. the number of completed batches/iterations, latency and throughput, with only a small penalty ($7\%$ to $9\%$) in model accuracy, a striking 56-fold (CNN) and 61-fold (VGG) model size compression and 2- to 4-fold communication cost reduction. FashionMNIST results are similar (see Table~\ref{tbl:Appendix_FashionMNIST_ModelsComparison} in Appendix~\ref{sec:Appendix_Additional_Evaluation}).
\section{Conclusion}
We introduce FedSparsify, an iterative process of federated pruning and tuning that produces highly sparse networks with learning performance similar to their dense counterparts. We use simple magnitude-based unstructured pruning in synchronous federated settings. Our scheme leads to a 15-fold model memory reduction, an average 4-fold model inference efficiency increase, and a 3-fold training communication cost decrease. In future work, we plan to explore more sophisticated magnitude-based model pruning and structured pruning strategies as well as pruning in asynchronous settings. Overall, FedSparsify serves as a strong baseline to evaluate future research in the Federated Learning Sparsification domain.

\bibliographystyle{unsrt}
\bibliography{bib/standardpruning.bib,bib/federatedpruning.bib,bib/any.bib}

\onecolumn
\appendix

\section{Proof of Thm.~\ref{thm:convergence}}
\label{appendix:proof}
We provide a brief proof sketch. We make the same assumptions as~\cite{jiang2022model,DBLP:conf/ijcai/HuGG21}, and they are as below. Refer Section~\ref{sec:FedSparsify} for notation.
\begin{assumption}\label{assn:1}
    Local objectives are smooth, i.e.,
        $
            \|\nabla f_k(w_1) - \nabla f_k(w_2)\| \leq
                L \|w_1-w_2\|,\  \forall \ w_1, w_2, k
        $ and some $L>0$.
\end{assumption}
\begin{assumption}\label{assn:2}
    Global objective is lipschitz, i.e.,
    $
        \|f(w_1) - f(w_2)\| \leq
            L_p \|w_1-w_2\|,\  \forall \ w_1, w_2
    $ and some $L_p>0$.
\end{assumption}
\begin{assumption}\label{assn:3}
    Client's stochastic gradients are unbiased, i.e.,
        $\E [g_k(w)] = \nabla f_k(w), \ \forall\ k, w$.
\end{assumption}
\begin{assumption}\label{assn:4}
    Local models have bounded gradient variance, i.e.,
        $\E \|g_k(w) - \nabla f_k(w)\|^2 \leq \sigma^2,  \  \forall\ k,w$.
\end{assumption}
\begin{assumption}\label{assn:5}
    The gradients from clients do not deviate much from the global model, i.e.,
    $\|\nabla f(w) - \nabla f_k(w)\|^2 \leq \epsilon^2, \ \forall\ k, w$.
\end{assumption}
\begin{assumption}\label{assn:6} 
    Time independent gradients, i.e.,
        $
            \E \left[g_k^{(t_1)} g_k^{(t_2)}\right]
                = \E \left[g_k^{(t_1)}\right]\  \E\left[ g_k^{(t_2)}\right],
                \ \forall\ t_1\neq t_2
        $.
\end{assumption}
\begin{assumption}\label{assn:7}
    Client independent gradients, i.e.,
        $
            \E \left[g_{k_1}^{(t_1)} g_{k_2}^{(t_2)}\right]
                = \mathrm{E} \left[g_{k_1}^{(t_1)}\right]                \E\left[ g_{k_2}^{(t_2)}\right],
                \ \forall\ k_1\neq k_2 $ and any $t_1, t_2
        $.
\end{assumption}

\begin{proof}
Since we enforce sparse structure found in previous iterations during client training and do not allow parameters to resurrect, we only need to demonstrate the convergence of the average over $\left\|\nabla f\left(w^{(t)}\right)\odot m^{(t)}\right\|$ terms.
Our proof technique is similar to previous approaches that have demonstrated convergence for federated learning under different scenarios~\cite{hu2022federated,jiang2022model,li2019convergence}.%
\\

\noindent Considering\
$ \E \left[f\left(w^{(t+1)}\right)\odot m^{(t)} - f\left(w^{(t)}\right)\right]$, due to $L$-smoothness assumption, we get ---
\begin{align}
    \E \left[f\left(w^{t+1} \odot m^{t}\right) - f\left(w^{t}\right)\right]
     &\leq
        \E \left\langle
            \nabla f\left(w^t\right), w^{t+1}\odot m^{t} - w^{t}
        \right\rangle
        + \frac L 2 \E
            \left\|w^{t+1} \odot m^t - w^{t}\right\|^2
    \label{eq:first}
\end{align}
\\

\noindent
Considering the first term from above,
\begin{align}
    \E &\left\langle
            \nabla f\left(w^t\right), w^{t+1}  \odot m^{t} - w^{t}
        \right\rangle \nonumber 
    \\
        &=\eta \E \left\langle
            \nabla f\left(w^t\right),
            - \frac 1 N \sum_{k=1}^N \sum_{i=0}^{S-1} g_k^{t,i}\odot m^t
        \right\rangle
        =\eta S \E \left\langle
            \nabla f\left(w^t\right),
            - \frac 1 {NS} \sum_{k=1}^{N} \sum_{i=0}^{S-1} g_k^{t,i}\odot m^t
        \right\rangle \nonumber
    \intertext{Due to independence assumptions~\ref{assn:6} and~\ref{assn:7}}
        & = \nonumber
        \eta S \left\langle
            {\nabla f\left(w^t\right)}\odot m^t,
            - \frac 1 {NS} \sum_{k=1}^N \sum_{i=0}^{S-1} {\nabla f_k\left(w^{t,s}_k\right)}\odot m^t
        \right\rangle
    \intertext{Since, $-2\langle a,b \rangle = -\|a\|^2 - \|b\|^2 + \|a-b\|^2$ }
        & = -\frac{\eta S} 2 \left\|{\nabla f\left(w^t\right)} \odot m^t\right\|^2
        - \frac{\eta S} 2 \left\|
            \frac 1 {NS} \sum_{k=1}^N\sum_{i=0}^{ S-1}
                {\nabla f_k\left(w^{t,i}_k\right)\odot m^t}
        \right\|^2\nonumber
        \\ \nonumber
        & \qquad \qquad \qquad+
        \frac{\eta S}{2} \left\|
            {\nabla f\left(w^t\right)} \odot m^t-
            \frac 1 {NS} \sum_{k=1}^N \sum_{i=0}^{ S-1}
                m^t\odot {\nabla  f_k\left(w^{t,i}_k\right)}
        \right\|^2 \nonumber
    \\
    \intertext{Due to $L$-smoothness assumption, and $\|\sum_{i=1}^{n} a_i\|^2 \leq n \sum_{i=1}^n\|a_i\|^2$}
        &\leq
        -\frac{\eta S}{2} \left\|m^t\odot {\nabla f\left(w^t\right)} \right\|^2
        - \frac{\eta S} 2 \left\|
            \frac 1 {NS} \sum_{k=1}^N\sum_{i=0}^{ S-1}
                {\nabla f_k\left(w^{t,i}_k\right)\odot m^t}
        \right\|^2
        \nonumber\\ 
        & \qquad \qquad \qquad+
        \frac {\eta S}{2}\frac {L^2} {N S} \sum_{k=1}^N \sum_{i=0}^{ S-1}
            \left\|{w^t}  - {w_k^{t,i}}\right\|^2
        \label{eq:dot_product_bound}
\end{align}
\\

\noindent
We know that for some independent random variables $[x_i]_{i=1}^P$, with $\E [x_i]  = \mu_i$, $E\left[\left(\sum x_i\right)^2\right] = \sum_i \mathrm{var}(x_i) + \left(\sum_i\mu_i\right)^2 $. Using this result and assumptions~\ref{assn:4}-\ref{assn:7}, for the second term in Eq.~\ref{eq:first}, we can establish that, 
\begin{align}
    \E \left\|w^{t+1} \odot m^t - w^{t}\right\|^2 
    &= \eta^2 \E\left\|
        \frac 1 N \sum_{k=1}^N \sum_{i=0}^{ S-1} m^t\odot {g_k^{t,i}}
    \right\|^2
    \\ &=
        S \sigma^2 \eta^2
        + \eta^2 \left\|\frac  1 N \sum_{k=1}^N \sum_{i=0}^{ S-1} 
            m^t\odot{\nabla f_k\left(w_k^{t,i}\right)}
        \right\|^2
        \label{eq:norm_of_gradients}
\end{align}
\\

\noindent
By repeating analysis similar to lemma 10 from~\cite{hu2022federated}, we can obtain the below result.
\begin{align}
\E\left\|{w_k^{t,i}}- {w^t} \right\|^2 \leq
        16\eta^2S^2 \left\|m^t \odot {\nabla f\left(w^{t}\right)}\right\|^2
        + 16\eta^2S^2\epsilon^2+4\eta^2S\sigma^2\label{eq:lem10}
\end{align}
\\

\noindent 
Substituting Eq.~\ref{eq:dot_product_bound},~\ref{eq:norm_of_gradients}, and~\ref{eq:lem10} in Eq.~\ref{eq:first}, we get
\begin{align}
    \nonumber
    \E & \left[  f\left(w^{t+1}\right)\odot m^{t} - f\left(w^{t}\right) \right]
    \\
        \leq 
        & -\frac{\eta S}{2} \left\|m^t\odot {\nabla f\left(w^t\right)} \right\|^2
        - \frac{\eta } {2S} \left\|
            \frac 1 {N} \sum_{k=1}^N\sum_{i=0}^{ S-1}
                {\nabla f_k\left(w^{t,i}_k\right)\odot m^t}
        \right\|^2
    \nonumber\\
    \nonumber
        & 
        +\frac {\eta L^2}{2N} \sum_{k=1}^N \sum_{i=0}^{ S-1}
            \left\|{w^t}  - {w_k^{t,i}}\right\|^2
        + \frac {L\eta^2}{2} \left( S \sigma^2
        + \left\|\frac  1 N \sum_{k=1}^N \sum_{i=0}^{ S-1} 
            m^t\odot{\nabla f_k\left(w_k^{t,i}\right)}
        \right\|^2\right)
    \\
    \nonumber
    \leq 
        & -\frac{\eta S}{2} \left\|m^t\odot {\nabla f\left(w^t\right)} \right\|^2
        +\left(\frac {L\eta^2}{2}- \frac{\eta} {2S}\right) \left\|
            \frac 1 {N} \sum_{k=1}^N\sum_{i=0}^{ S-1}
                {\nabla f_k\left(w^{t,i}_k\right)\odot m^t}
        \right\|^2
    \\
    \nonumber
        & 
        +\frac {\eta L^2}{2N} NS 
            \left(16\eta^2S^2 \left\|m^t \odot {\nabla f\left(w^{t}\right)} \right\|^2
                + 16\eta^2S^2\epsilon^2+4\eta^2S\sigma^2
            \right)
        + \frac {LS \sigma^2\eta^2}{2} 
    \\
\intertext{Since, $\eta < (4\sqrt 2LS)^{-1}$, then $\left(\frac {L\eta^2}{2}- \frac{\eta} {2S}\right) < 0$}
    \leq
        & \frac{\eta S}{2} \left(-1 + 16L^2\eta^2 S^2\right)
        \left\| m^{t}\odot {\nabla f(w^{t})} \right\|^2
        +\left(\frac {\eta^2LS}{2}+2L^2\eta^3 S^2\right)\sigma^2
        +8L^2\eta^3 S^3\epsilon^2    
    \label{eq:weight_diff_bound}
\end{align}

\noindent 
Above result establishes bound for the weight updates during federated training round. However, pruning further changes the model outputs, but its effect can be bounded due to the lipschitz assumption (Assumption~\ref{assn:2}).
\begin{align}
E \biggl[f\left(w^{t+1}\right)- f\left(w^{t+1}\right)\odot m^{t} \biggr]
    \leq 
    L_p\left\|w^{t+1}- w^{t+1}\odot  m^{t}\right\|
    \label{eq:prune_diff}
\end{align}
\\

\noindent 
Adding Eq.~\ref{eq:weight_diff_bound} and~\ref{eq:prune_diff}, we get ---

\begin{align*}
\E \left[f\left(w^{t+1}\right) - f\left(w^{t}\right)\right] \leq 
    & \frac{\eta S}{2} \left(-1 + 16L^2\eta^2 S^2\right)
        \left\| m^{t}\odot {\nabla f(w^{t})} \right\|^2
        + \left(\frac {\eta^2LS}{2}+2L^2\eta^3 S^2\right)\sigma^2
    \\
    & +8L^2\eta^3 S^3\epsilon^2   + L_p\left\|w^{t+1}- w^{t+1}\odot  m^{t}\right\| 
\end{align*}
\begin{align*}
\intertext{Since, $\eta \leq (4\sqrt 2 LS)^{-1} \implies \left(1 - 16L^2\eta^2 S^2\right) \leq 1/2$ }
\frac{\eta S}{2} \left(1 - 16L^2\eta^2 S^2\right)
    \left\| m^{t}\odot {\nabla f(w^{t})} \right\|^2  \leq
    & \E \left[f\left(w^{t}\right) - f\left(w^{t+1}\right)\right]   
        +\left(\frac {\eta^2LS}{2}+2L^2\eta^3 S^2\right)\sigma^2
    \\
    & +8L^2\eta^3 S^3\epsilon^2   
    + L_p\left\|w^{t+1}- w^{t+1}\odot  m^{t}\right\|    
    \\
    \frac{\eta S}{4}
    \left\| m^{t}\odot {\nabla f(w^{t})} \right\|^2  \leq & \E \left[f\left(w^{t}\right) - f\left(w^{t+1}\right)\right]   
        +\left(\frac {\eta^2LS}{2}+2L^2\eta^3 S^2\right)\sigma^2
    \\
\end{align*}

\noindent
Equivalently,
\begin{align*}
\left\| m^{t}\odot {\nabla f(w^{t})} \right\|^2  &\leq
    \frac 4 {\eta S} \E \left[f\left(w^{t}\right) - f\left(w^{t+1}\right)\right]
    + \frac 4 {\eta S}  L_p\left\|w^{t+1}- w^{t+1}\odot  m^{t}\right\|
    \\
    &+ 2{\eta L} \left[\left(1+4L\eta S\right)\sigma^2 + 16L\eta S^2\epsilon^2\right]
\end{align*}

\noindent
Summing over all the time steps, and noting that
\[
\E \left[f\left(w^{1}\right) - f\left(w^{t+1}\right)\right] \geq
        \E \left[f\left(w^{1}\right) - f\left(w^{*}\right)\right]
\]
gives the desired result. 
\end{proof}

\section{Federated Experiments Specifications}\label{sec:Appendix_FederatedExperimentsSpecifications}

In this section, we describe federated environments and the hyperparameter configurations for all the experiments. 
The random seed for all the experiments was set to 1990. All experiments were run on a dedicated GPU server equipped with 4~Quadro RTX 6000/8000 graphics cards of 50~GB RAM each, 31~Intel(R) Xeon(R) Gold 5217 CPU @ 3.00~GHz, and 251~GB DDR4 RAM.

\begin{figure}[htpb]
  \centering
  \subfloat[10 Clients IID]{\includegraphics[width=0.21\linewidth]{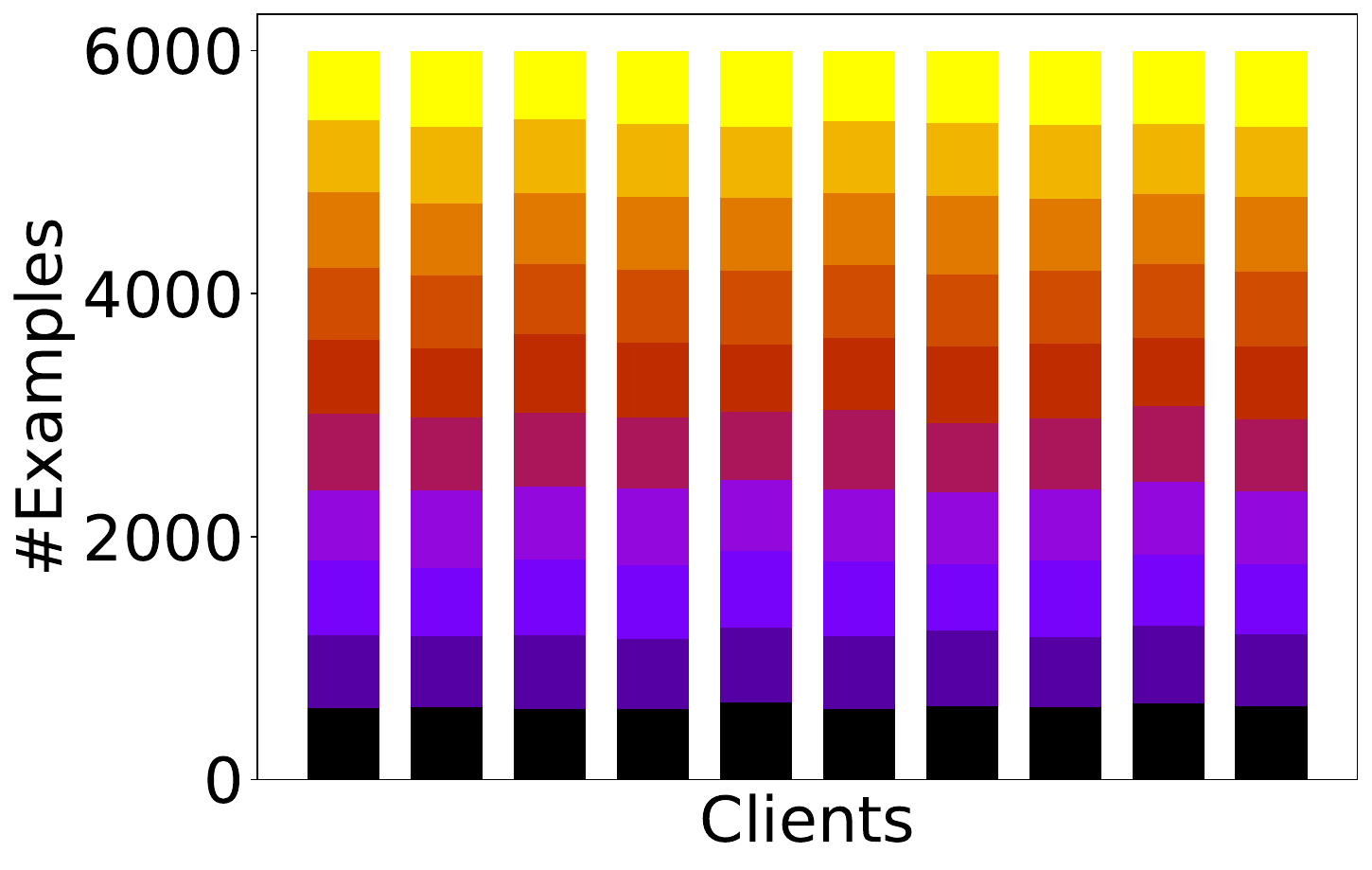}\label{subfig:Appendix_FashionMMNIST_DataDistribution_Clients10_IID}}\quad
  \subfloat[100 Clients IID]{\includegraphics[width=0.21\linewidth]{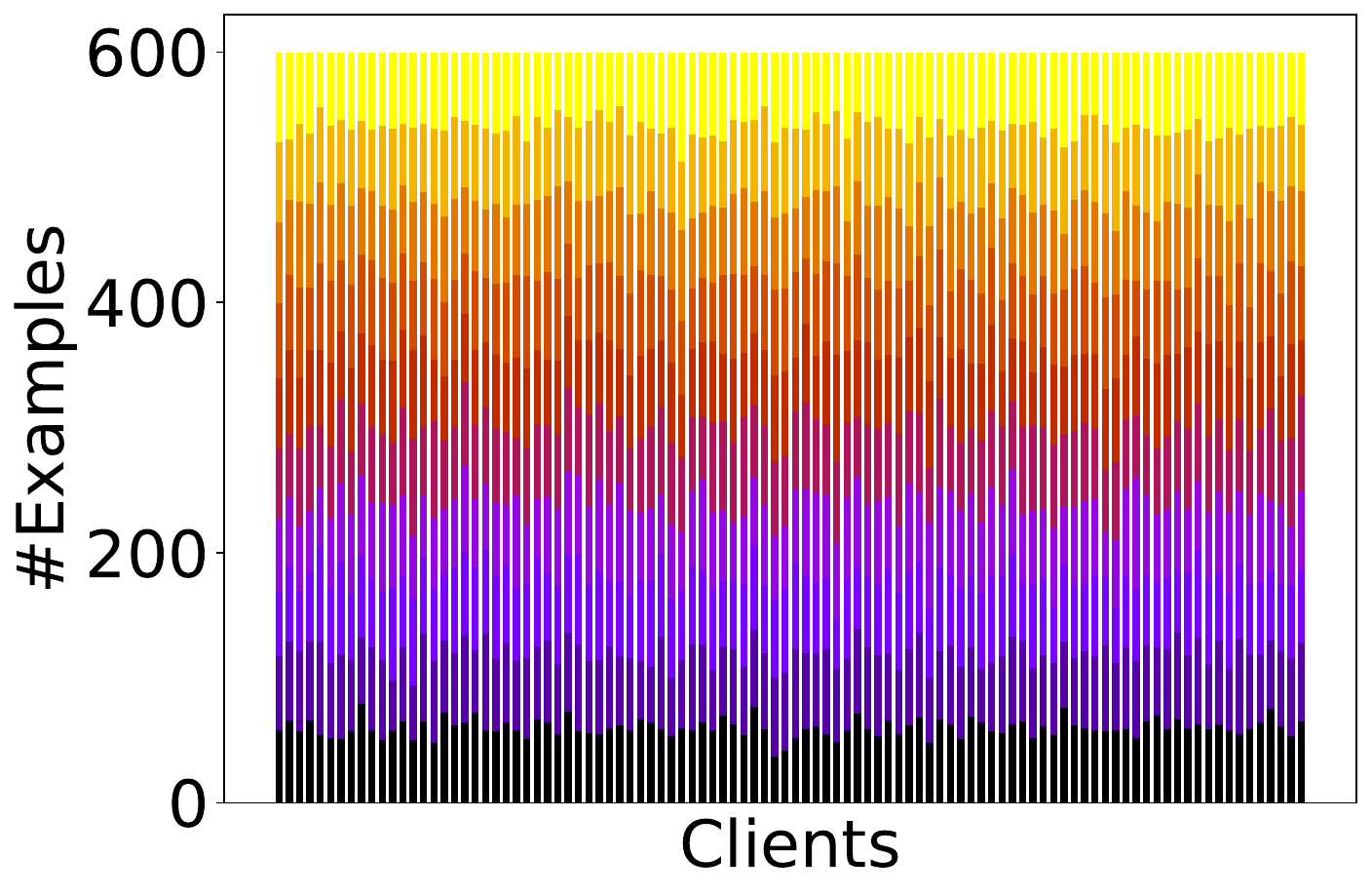}\label{subfig:Appendix_FashionMMNIST_DataDistribution_Clients100_IID}}\quad
  \subfloat[10 Clients Non-IID(2)]{\includegraphics[width=0.21\linewidth]{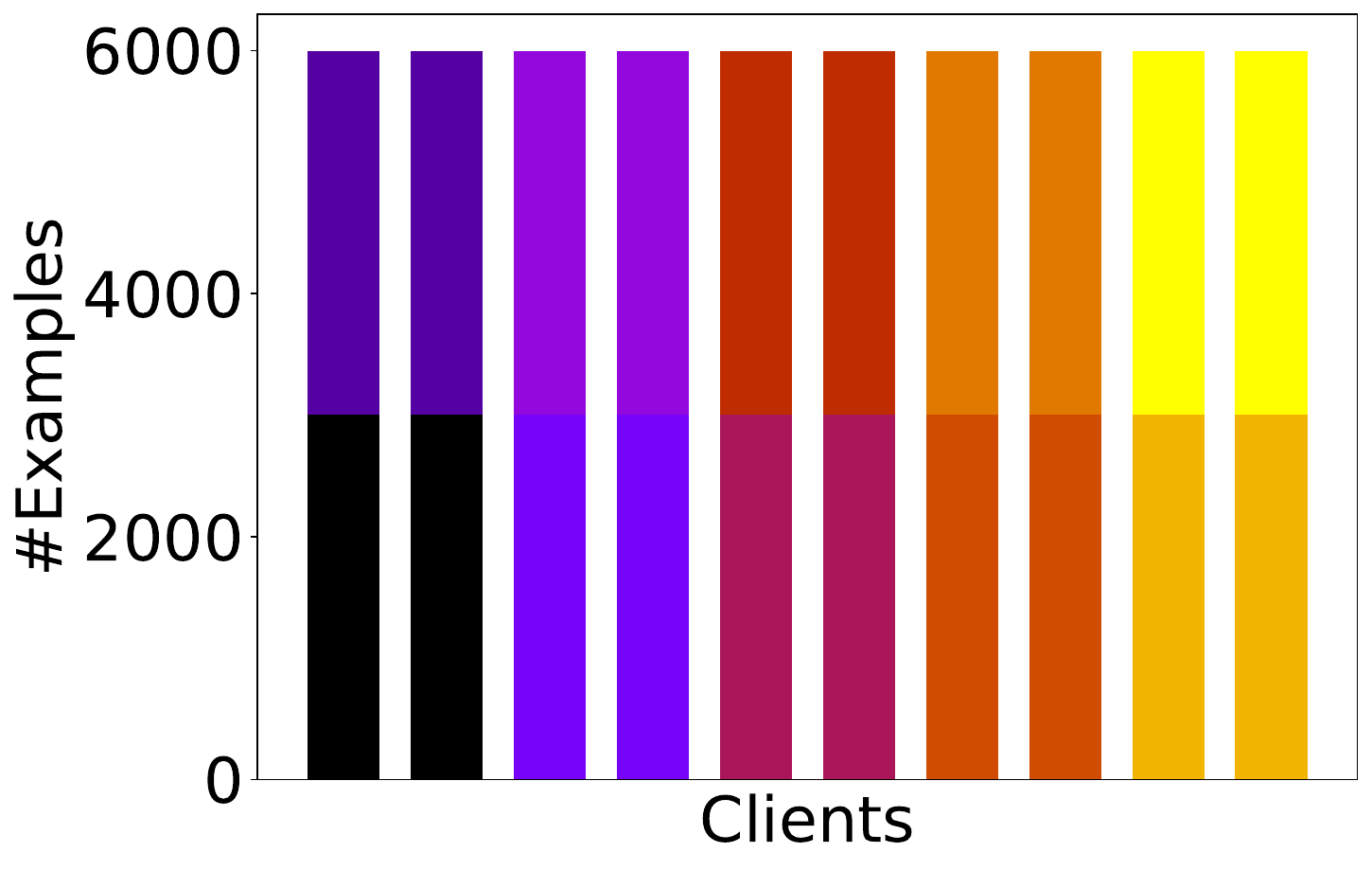}\label{subfig:Appendix_FashionMMNIST_DataDistribution_Clients10_NonIID}}\quad
  \subfloat[100 Clients Non-IID(2)]{\includegraphics[width=0.21\linewidth]{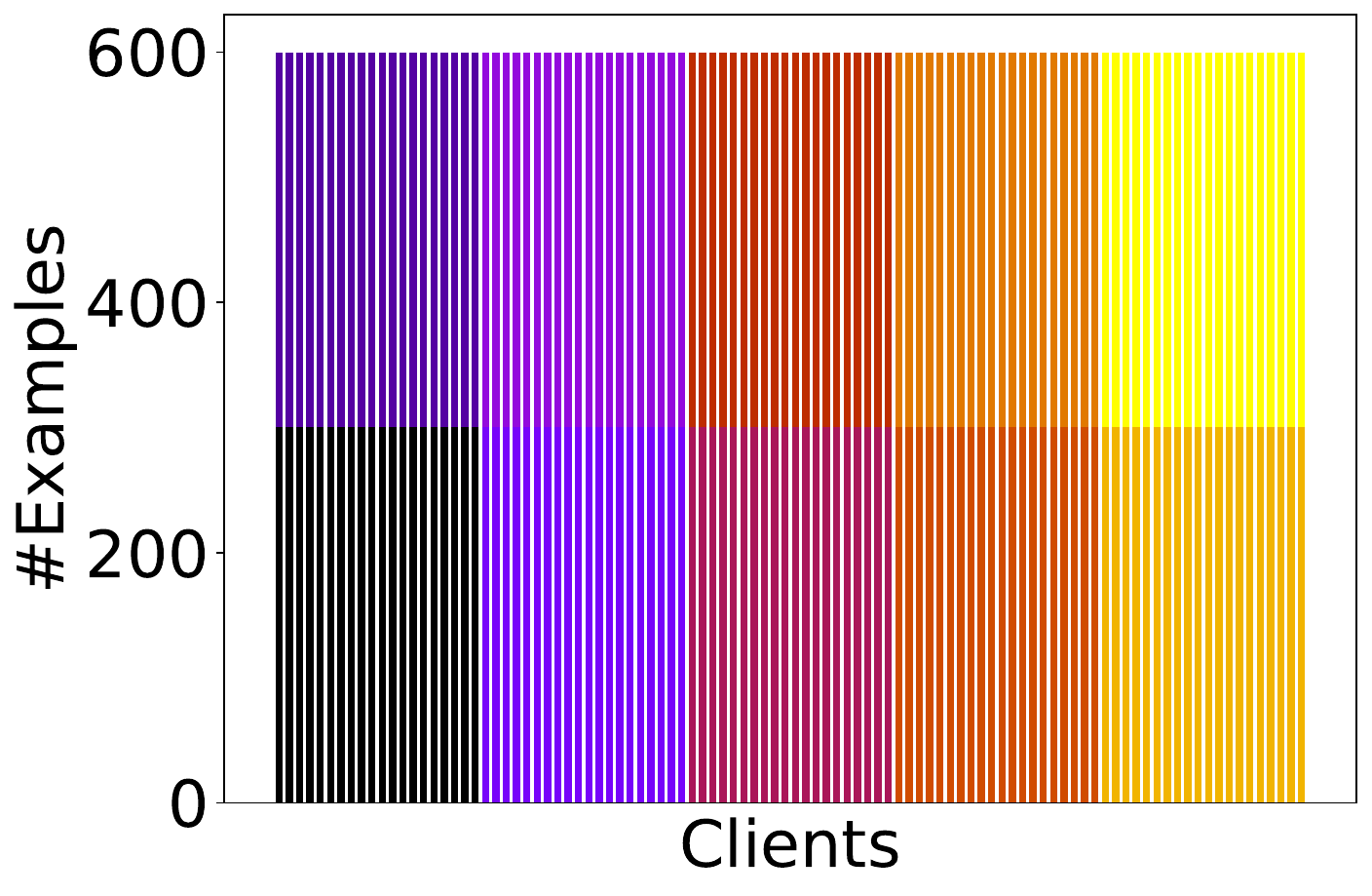}\label{subfig:Appendix_FashionMMNIST_DataDistribution_Clients100_NonIID}}
  \caption{FashionMNIST Federated Data Distributions.}
  \label{fig:Appendix_FashionMMNIST_DataDistribution}
\end{figure}

\begin{figure}[htpb]
  \centering
  \subfloat[10 Clients IID]{\includegraphics[width=0.21\linewidth]{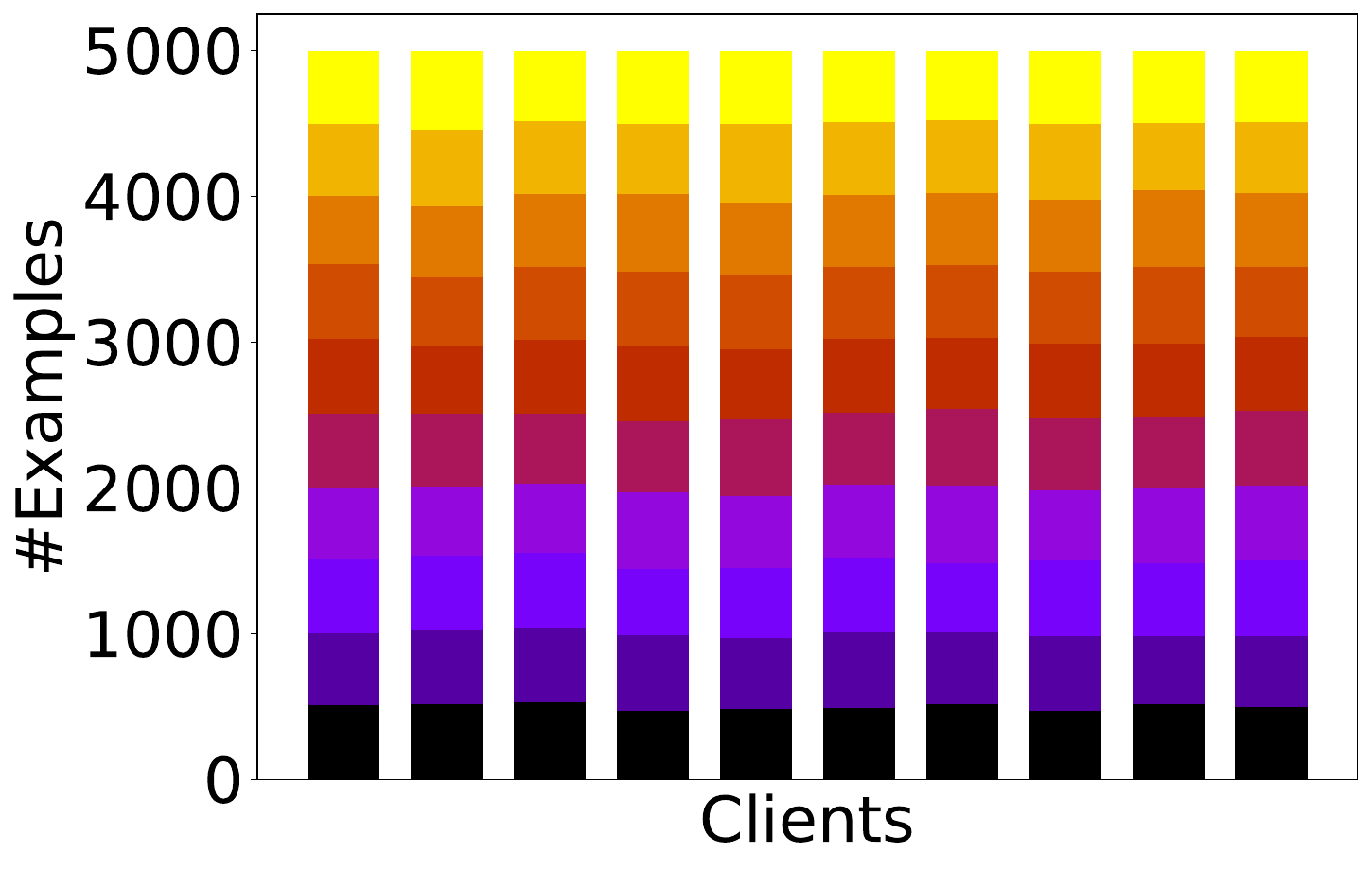}\label{subfig:Appendix_CIFAR10_DataDistribution_Clients10_IID}}\quad
  \subfloat[100 Clients IID]{\includegraphics[width=0.21\linewidth]{figures/appendix/cifar10_datadistribution_clients10_iid.pdf}\label{subfig:Appendix_CIFAR10_DataDistribution_Clients100_IID}}\quad
  \subfloat[10 Clients Non-IID(5)]{\includegraphics[width=0.21\linewidth]{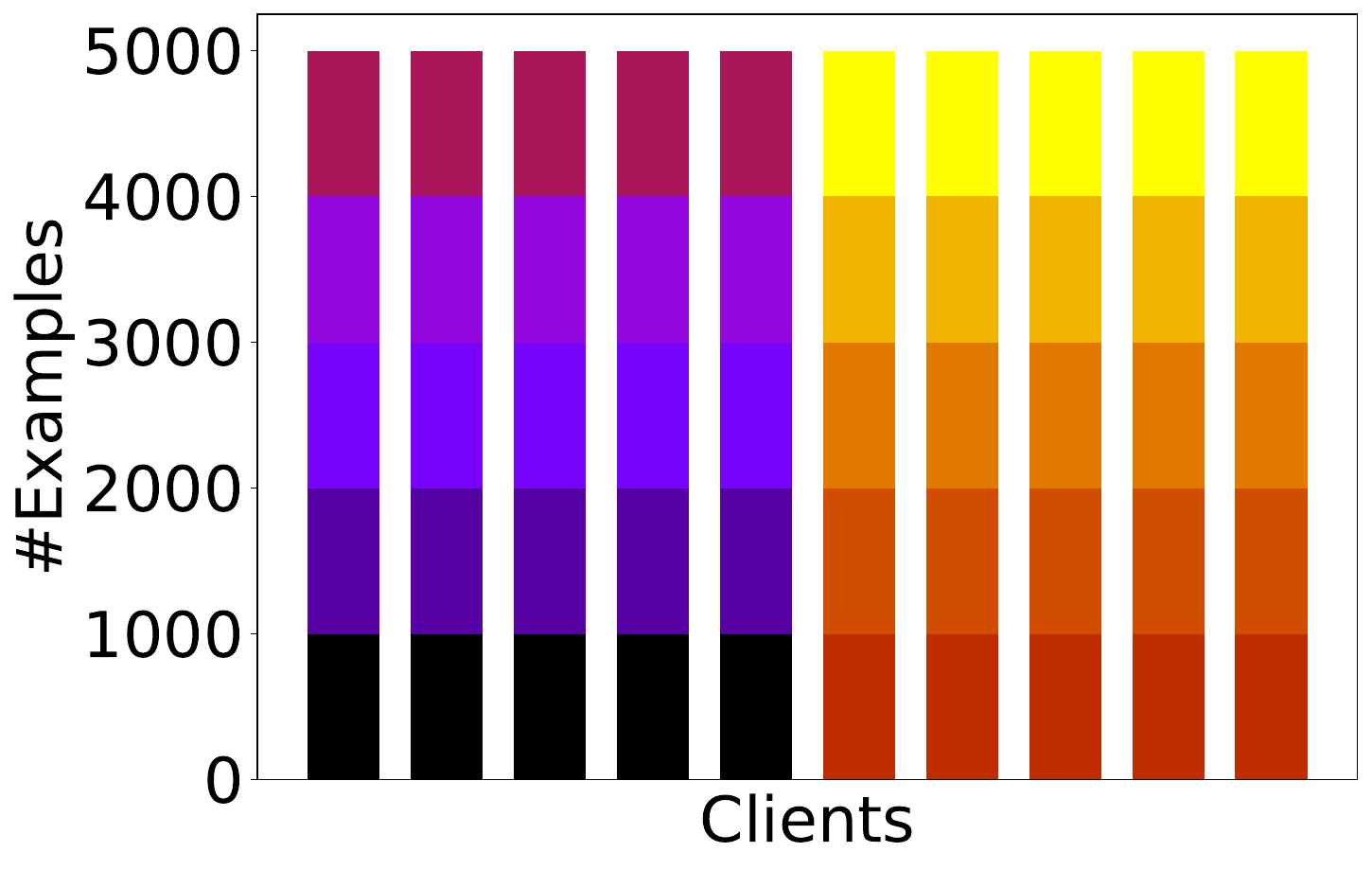}\label{subfig:Appendix_CIFAR10_DataDistribution_Clients10_NonIID}}\quad
  \subfloat[100 Clients Non-IID(5)]{\includegraphics[width=0.21\linewidth]{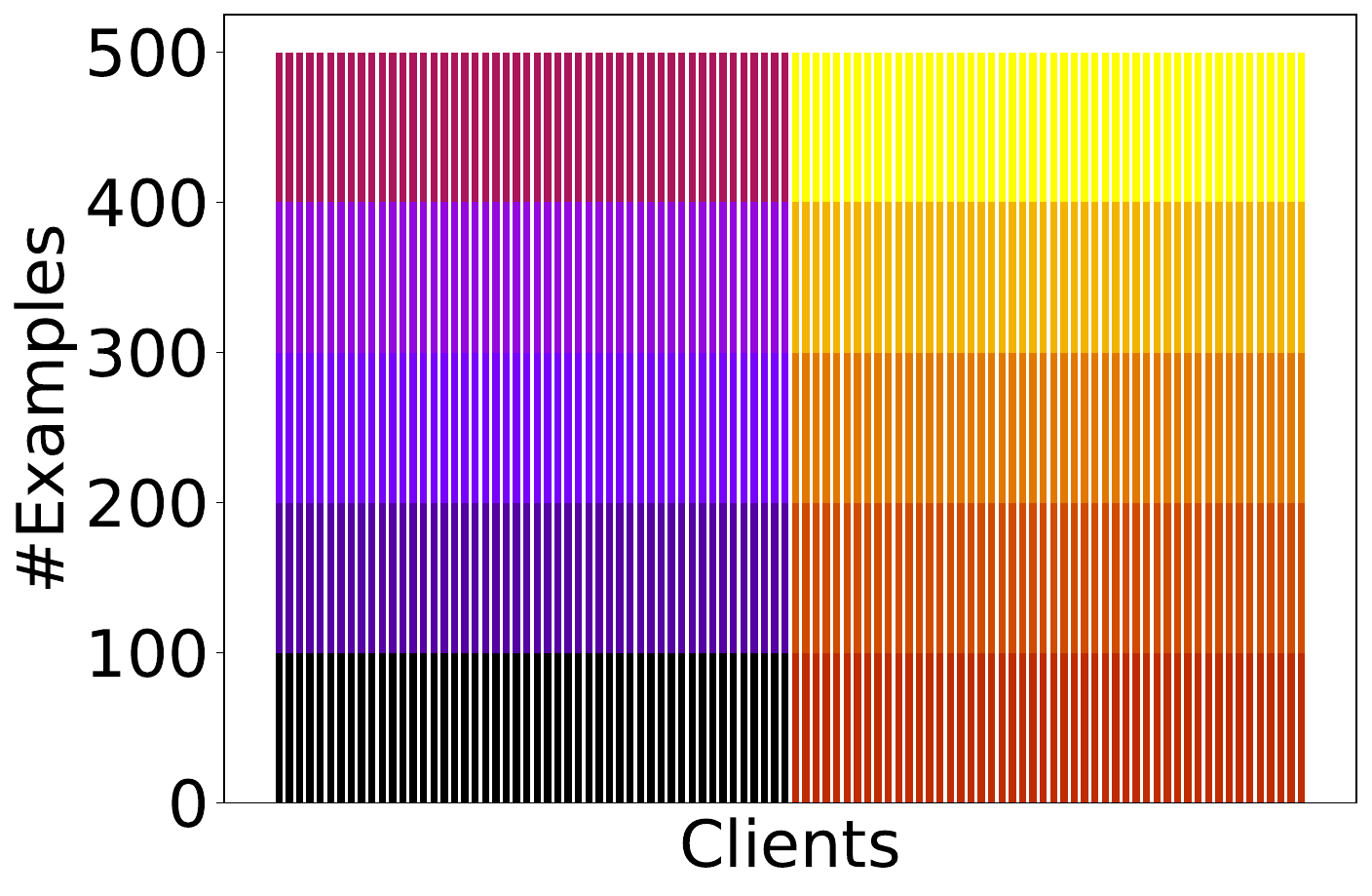}\label{subfig:Appendix_CIFAR10_DataDistribution_Clients100_NonIID}}
  \caption{CIFAR-10 Federated Data Distributions}
  \label{fig:Appendix_CIFAR10_DataDistribution}
\end{figure}

\begin{figure}[htpb]
  \centering
  \subfloat[10 Clients IID]{\includegraphics[width=0.21\linewidth]{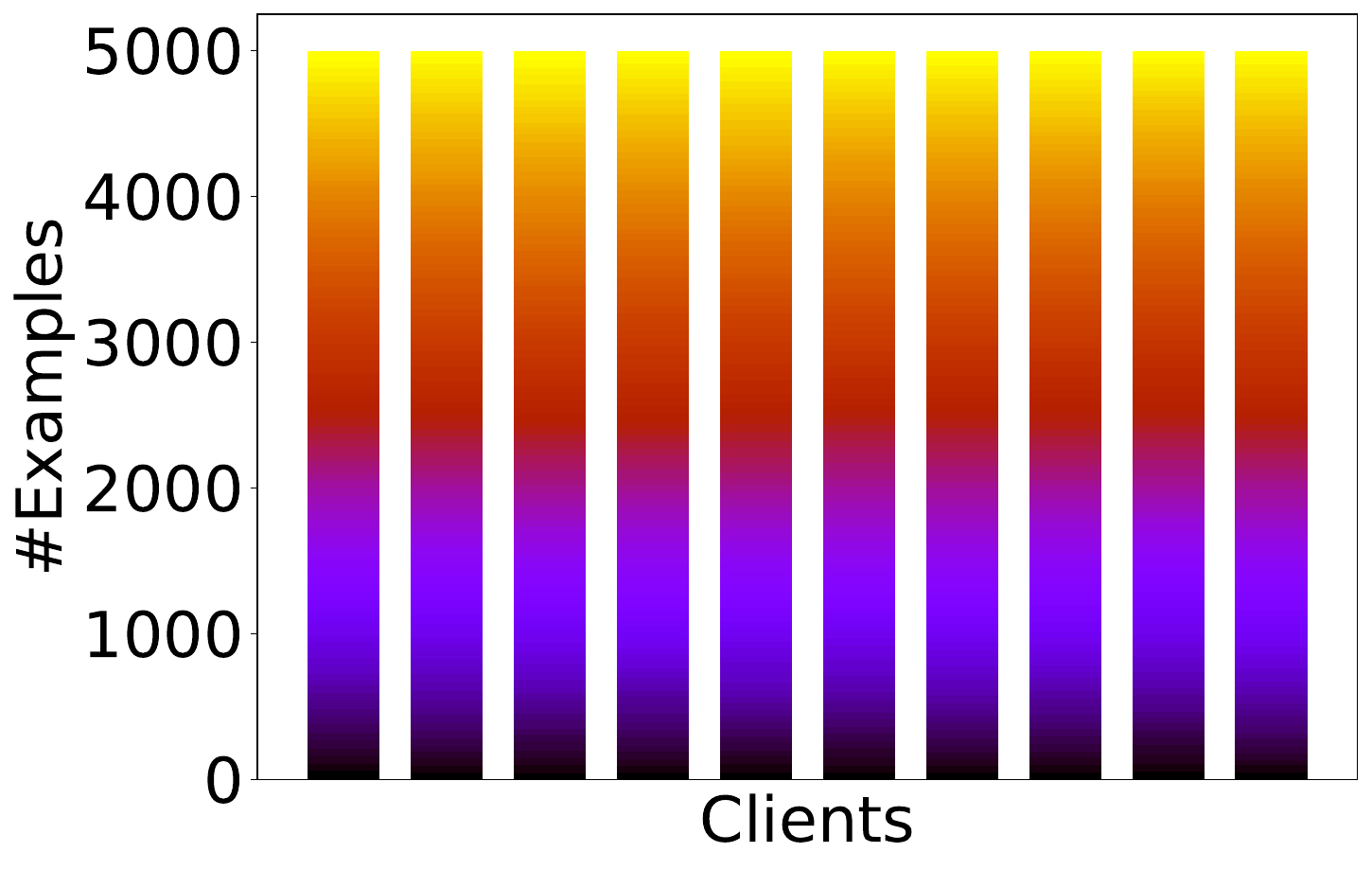}\label{subfig:Appendix_CIFAR100_DataDistribution_Clients10_IID}}\quad
  \subfloat[100 Clients IID]{\includegraphics[width=0.21\linewidth]{figures/appendix/cifar100_datadistribution_clients10_iid.pdf}\label{subfig:Appendix_CIFAR100_DataDistribution_Clients100_IID}}\quad
  \subfloat[10 Clients Non-IID(50)]{\includegraphics[width=0.21\linewidth]{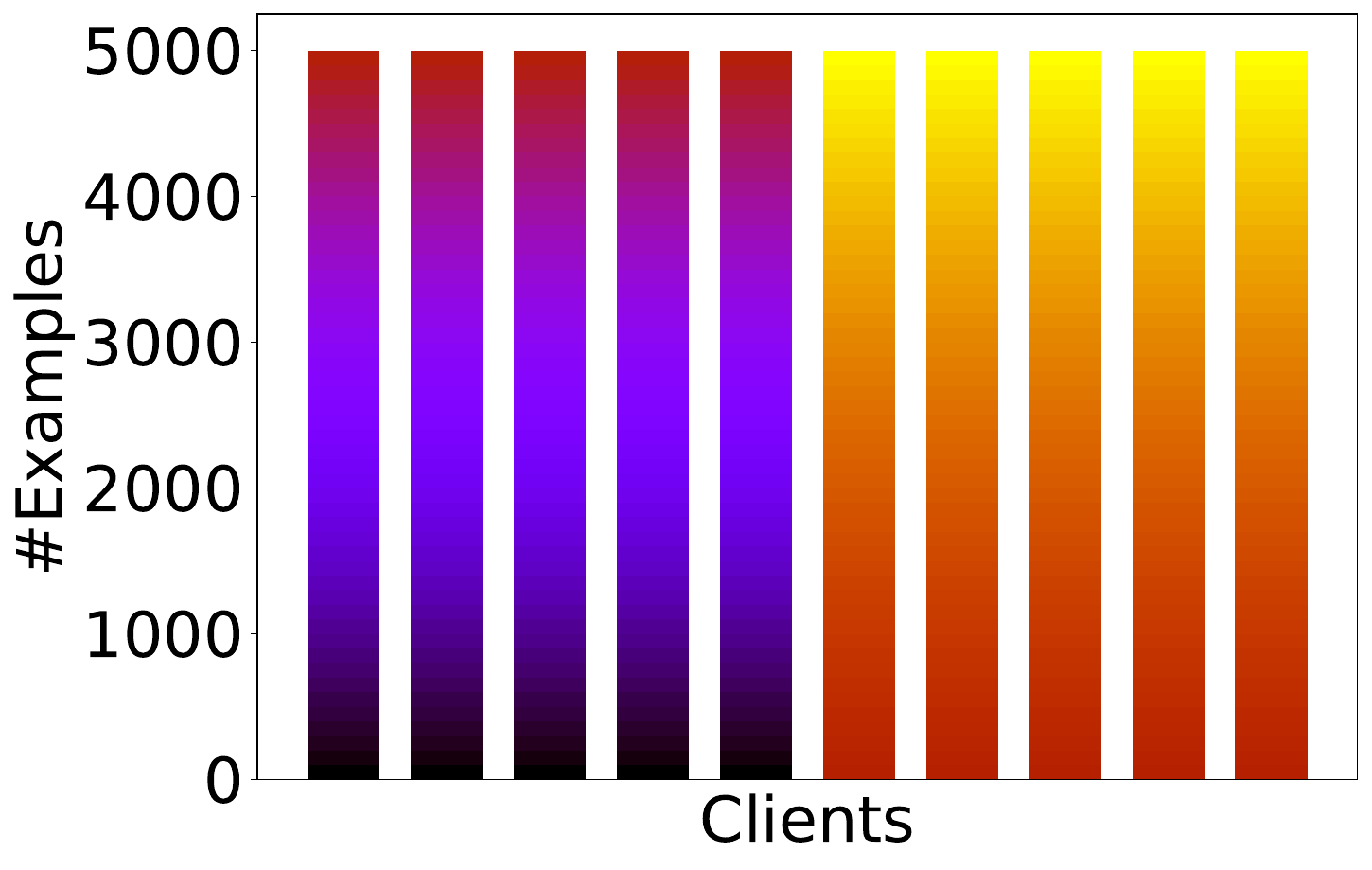}\label{subfig:Appendix_CIFAR100_DataDistribution_Clients10_NonIID}}\quad
  \subfloat[100 Clients Non-IID(50)]{\includegraphics[width=0.21\linewidth]{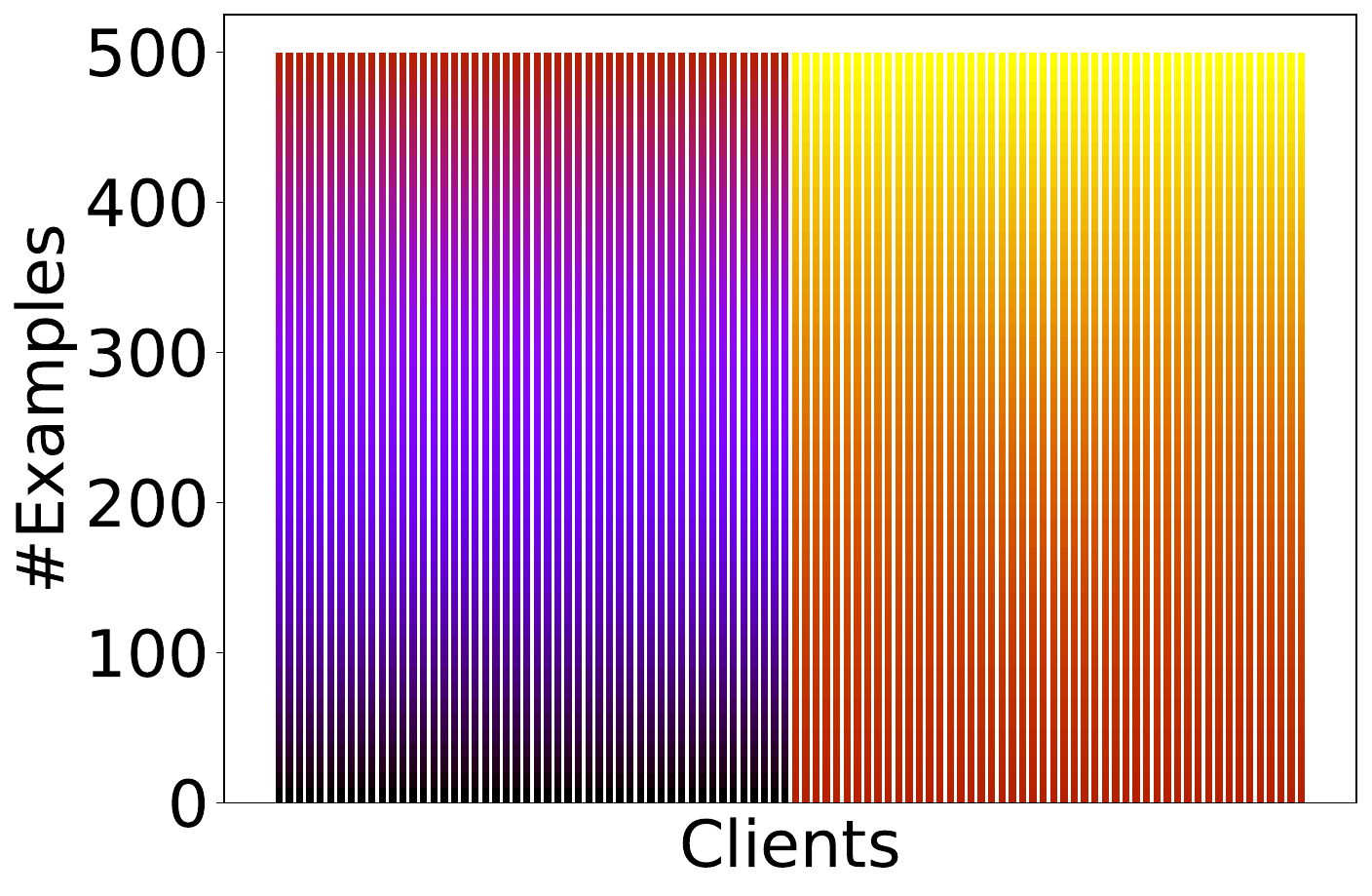}\label{subfig:Appendix_CIFAR100_DataDistribution_Clients100_NonIID}}
  \caption{CIFAR-100 Federated Data Distributions}
  \label{fig:Appendix_CIFAR100_DataDistribution}
\end{figure}

\paragraph{Federated Data Distributions.} Figures~\ref{fig:Appendix_FashionMMNIST_DataDistribution},~\ref{fig:Appendix_CIFAR10_DataDistribution}, and~\ref{fig:Appendix_CIFAR100_DataDistribution} present the federated data distributions used in this work for FashionMNIST, CIFAR-10, and CIFAR 100 datasets, respectively. In all the figures, the x-axis refers to the clients, with one bar plot per client, and the y-axis shows the number of samples per client and per class. To represent the class distribution inside each client, we use sequential coloring for the bar plot with each increasing color representing a different class.

\paragraph{Federated Models Hyperparameters.} 
Every federated model for FashionMNIST and CIFAR-10 was trained for 200 rounds, and 100 rounds for CIFAR-100. Each client trains for 4 local epochs in one round. We used batch size  32 for FashionMNIST and CIFAR-10 and batch size 128 for CIFAR-100. For CIFAR-100 with 5000 and 500 examples per client in the environment of 10 and 100 clients, respectively, this translates to $4*5000/128=156$ and $4*500/128=15$ local steps per client per round. We used SGD with learning rate 0.02 for FashionMNIST, and SGD with momentum = 0.75 and learning rate  0.005 for CIFAR-10. For FedProx, the proximal term $\mu$ is kept constant at 0.001. 
For FedProx and CIFAR-100, we searched for a proximal term in  the range 0.01, 0.001, and 0.0001, with 0.001 performing the best. For all the other CIFAR-100 experiments, we used SGD with momentum = 0.9 and learning rate 0.01. 

\paragraph{FedSparsify Hyperparameters.} 
For all \textit{FedSparsify-Local} and \textit{FedSparsify-Global} experiments, sparsification starts at round 1 ($t_0 = 1$), initial degree of sparsification is 0 ($S_0 = 0$), sparsification frequency is 1 ($F=1$, 1 round of tuning), and the exponent value is 3 ($n=3$).  

\begin{figure}[htpb]
  \centering
  \subfloat[Sparsification frequency hyperparameter exploration with respect to federation rounds (x-axis). Left y-axis and solid lines show accuracy, right y-axis show global model parameters progression. The higher the sparsification frequency, $F$, the bigger the drop in model performance.]{
  \includegraphics[width=0.45\linewidth]{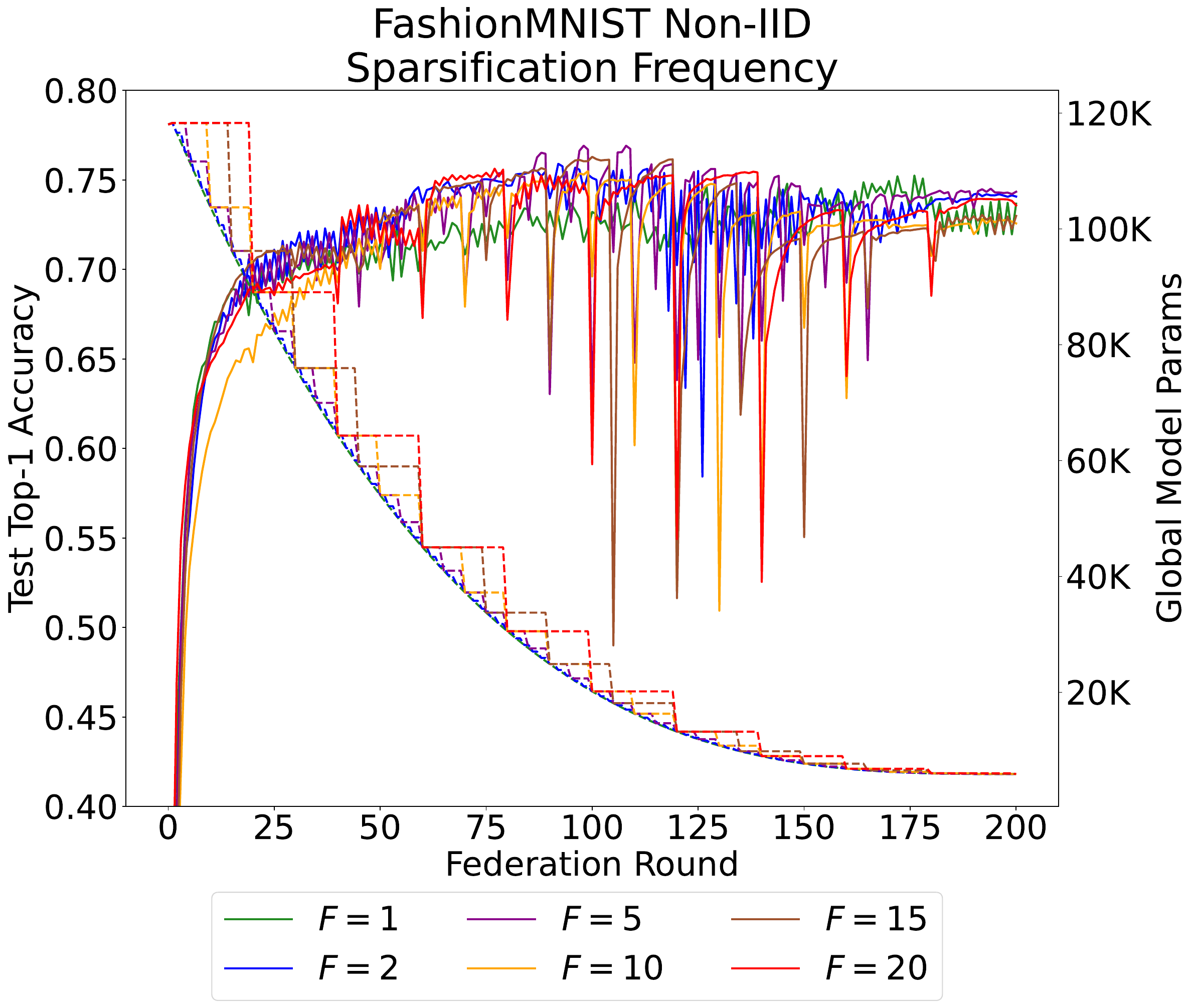}\label{subfig:Appendix_FashionMNIST_SparsificationFrequencyExploration_FederationRound}}\quad
  \subfloat[Sparsification frequency hyperparameter exploration with respect to transmission cost (x-axis). The higher the sparsifcation frequency is the higher the incurred communication cost is during training.]{\includegraphics[width=0.45\linewidth]{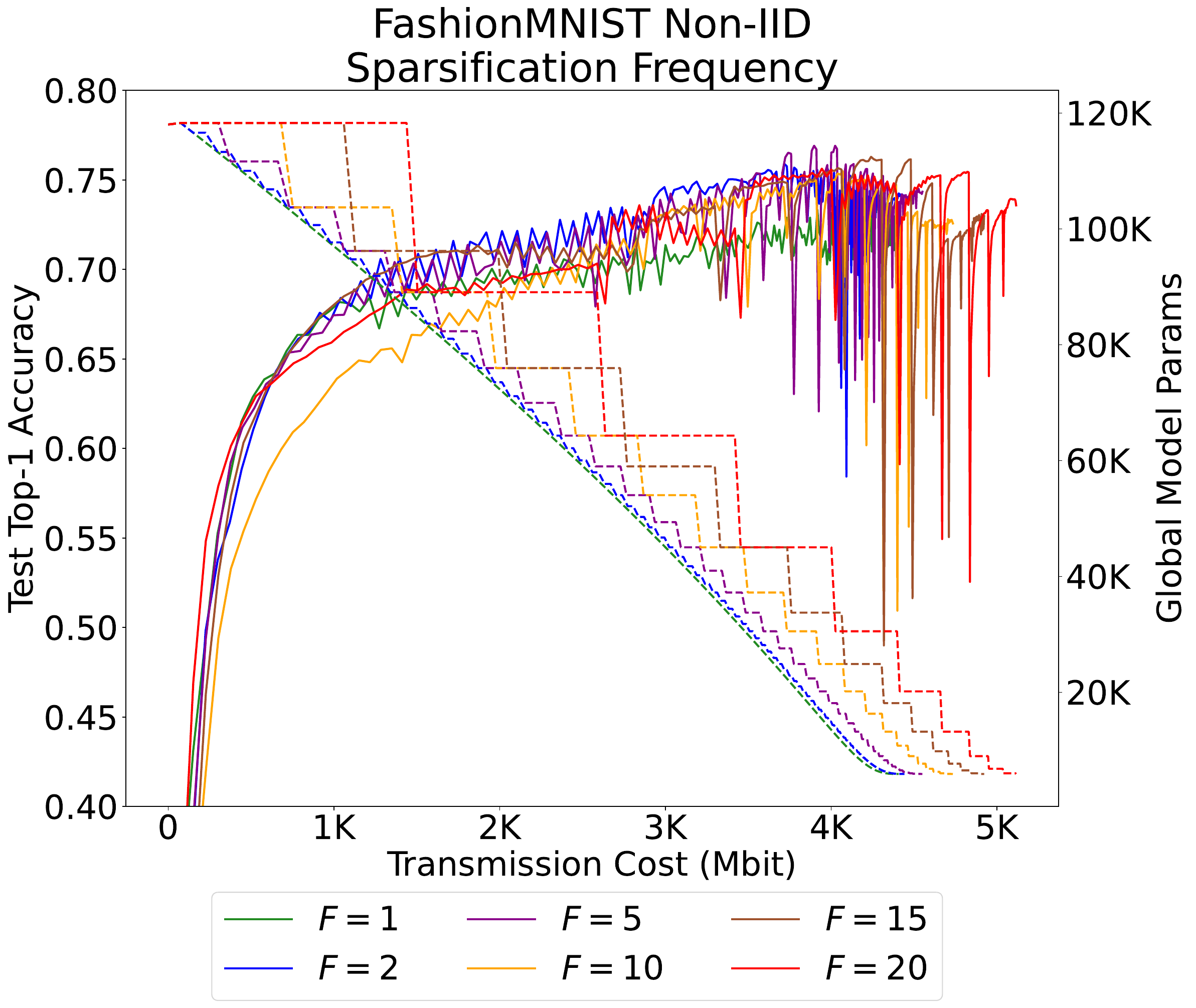}\label{subfig:Appendix_FashionMNIST_SparsificationFrequencyExploration_CommunicationCost}}
  
  \subfloat[Exponent hyperparameter exploration with respect to federation rounds (x-axis). Left y-axis and solid lines show accuracy, right y-axis show global model parameters progression. The higher the exponent value is (e.g., $n=6, 12$), the greater the number of pruning weights is during early training stages.]{\includegraphics[width=0.45\linewidth]{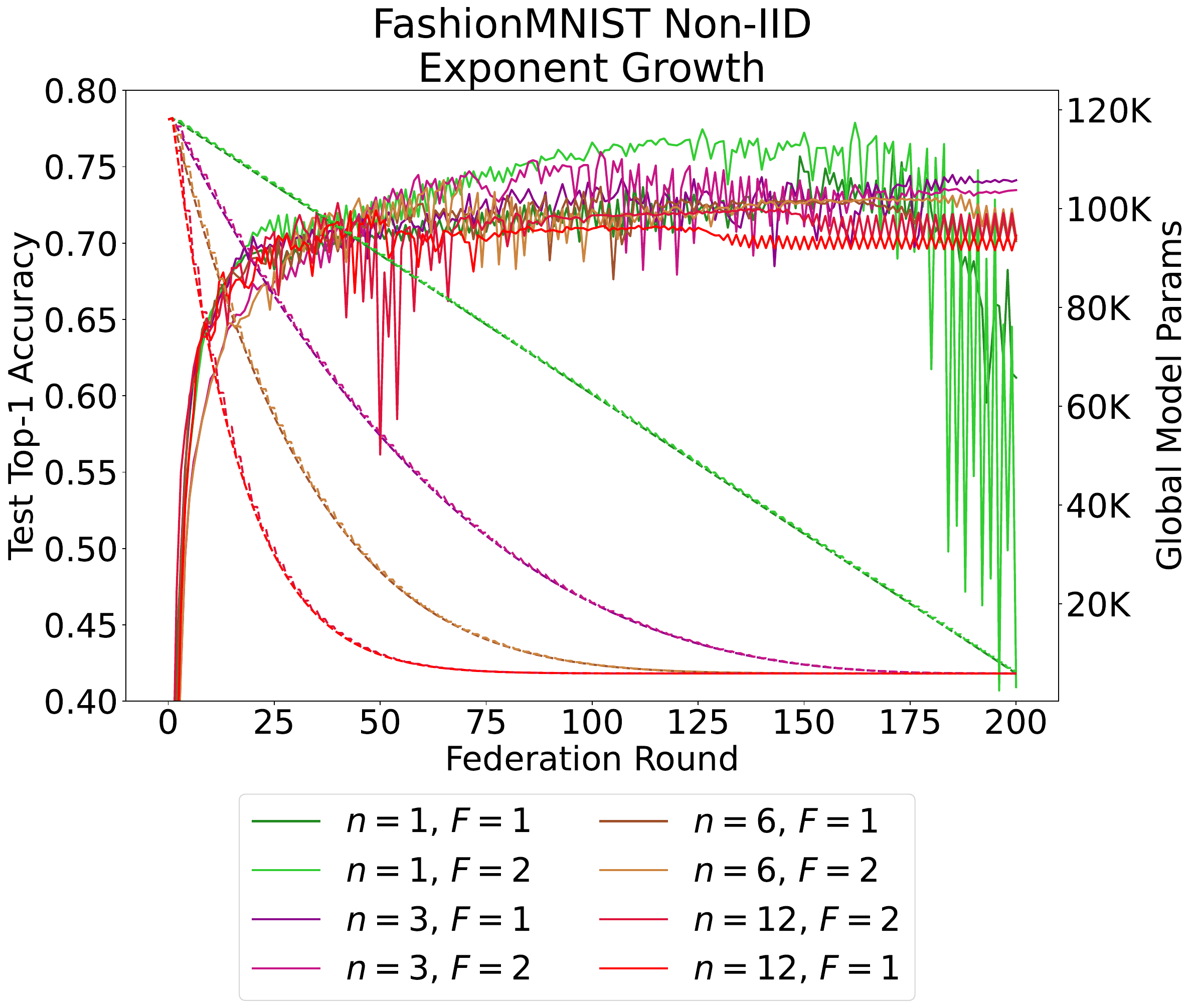}\label{subfig:Appendix_FedSparsify_Exponent_FederationRound}}\quad
  \subfloat[Exponent hyperparameter exploration with respect to transmission cost (x-axis). Left y-axis and solid lines show accuracy, right y-axis show global model parameters progression. The higher the exponent value is (e.g., $n=6, 12$), the smaller the transmission cost is, but with worse final performance.]{\includegraphics[width=0.45\linewidth]{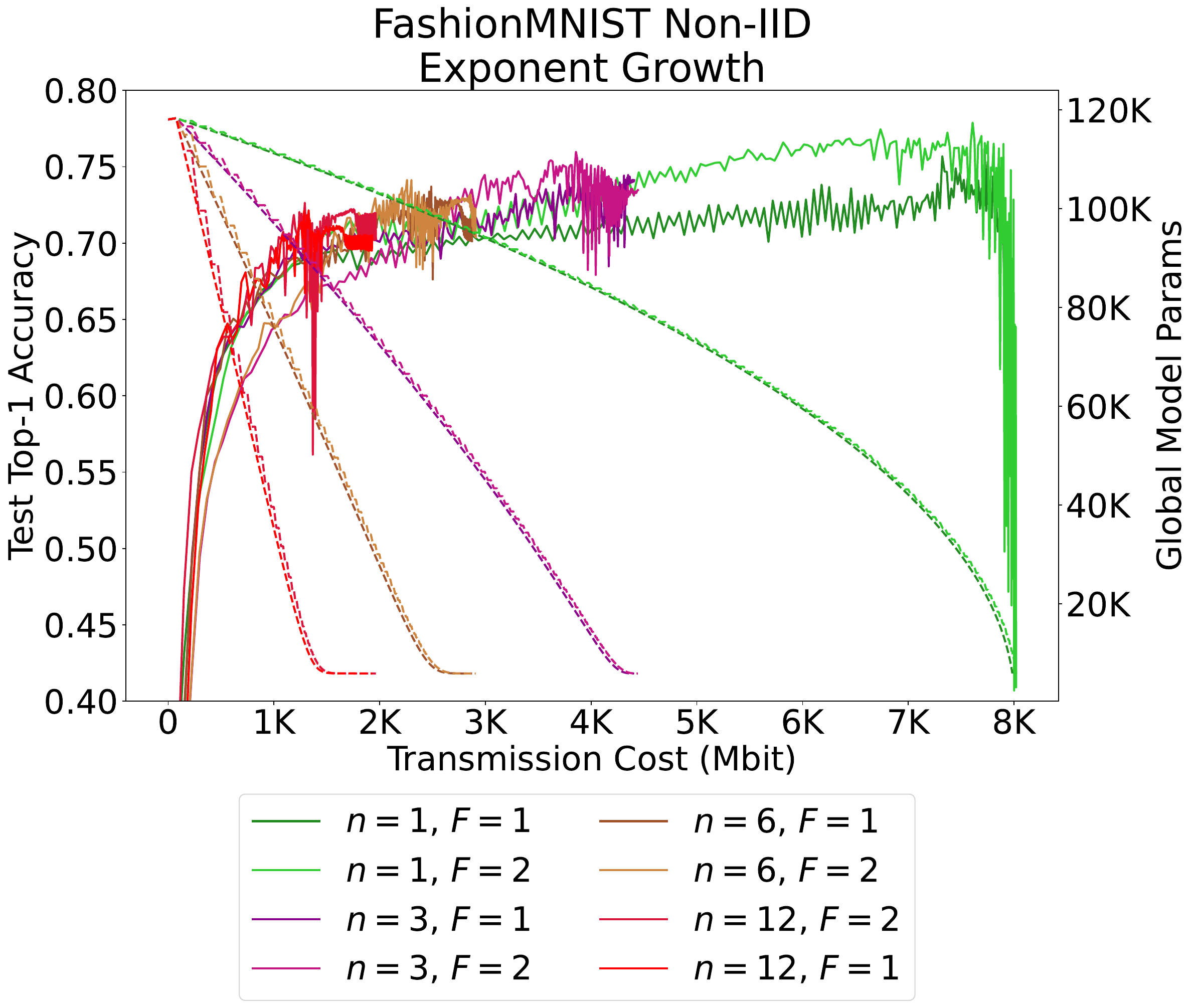}\label{subfig:Appendix_FedSparsify_Exponent_CommunicationCost}}
  
  \caption{FedSparify Tuning. Top row shows the convergence when exploring different sparsification frequency values with FedSparsify-Global at 0.95 sparsity on FashionMNIST with 10 clients over Non-IID data distribution with respect to Federation Rounds (Figure~\ref{subfig:Appendix_FashionMNIST_SparsificationFrequencyExploration_FederationRound}) and Transmission Cost (Figure~\ref{subfig:Appendix_FashionMNIST_SparsificationFrequencyExploration_CommunicationCost}); exponent value for these experiments is set to 3. The bottom row Figures~\ref{subfig:Appendix_FedSparsify_Exponent_FederationRound} and~\ref{subfig:Appendix_FedSparsify_Exponent_CommunicationCost} show the exponent hyperparameter exploration in terms of Federation Rounds and Transmission Cost, respectively.}
  \label{fig:Appendix_FedSparsify_Tuning}
\end{figure}

\paragraph{FedSparsify Tuning.}
During frequency value exploration, we observed that frequency values of $F=1$ and $F=2$ behave similarly. However, for higher values of frequency (e.g., $F\in\{5, 10, 15, 20\}$), i.e., more rounds of fine-tuning, there is a big drop in the model performance when pruning takes place since a larger ratio of weights is pruned in a single pruning step. We show the effects of training with different pruning frequencies in terms of Federation Rounds in Figure~\ref{subfig:Appendix_FashionMNIST_SparsificationFrequencyExploration_FederationRound} and in terms of Transmission Cost in Figure~\ref{subfig:Appendix_FashionMNIST_SparsificationFrequencyExploration_CommunicationCost}; in the next section~\ref{sec:Appendix_Additional_Evaluation} we discuss how Transmission cost is measured exactly. In Figures~\ref{subfig:Appendix_FedSparsify_Exponent_FederationRound} and~\ref{subfig:Appendix_FedSparsify_Exponent_CommunicationCost} we show the effects of different exponent values in the final performance. As expected, for high exponent values the pruning effect is more profound and leads to pruned models with reduced final performance. This is in contrast to the very small exponent value (exp=1) that can learn better models mid-training due to the larger number of available non-pruned weights but fails to retain that performance at the final rounds, where pruning is more aggressive. An exponent value equal to 3 provides a good trade-off between sparsification and model performance for our scheme. Similar effects hold in the case of transmission cost, where high exponent values lead to reduced transmission costs but degraded performance, and small exponent values higher transmission cost and better performance. Still, an exponent value of 3 provides a good alternative to balance transmission cost and learning performance. Finally, for \textit{FedSparsify-Local}, we use Majority Voting as the aggregation rule of the local models, while for Random pruning baseline and \textit{FedSparsify-Global}, we use FedAvg's aggregation rule.

\begin{figure}[htpb]
  \centering

  \subfloat[Learners prune their local model right after local training is complete. Thereafter, they share their pruned model along with the sparsification binary mask with the controller (server) and the controller aggregates the pruned models using Majority Voting.]{\includegraphics[width=0.35\textwidth]{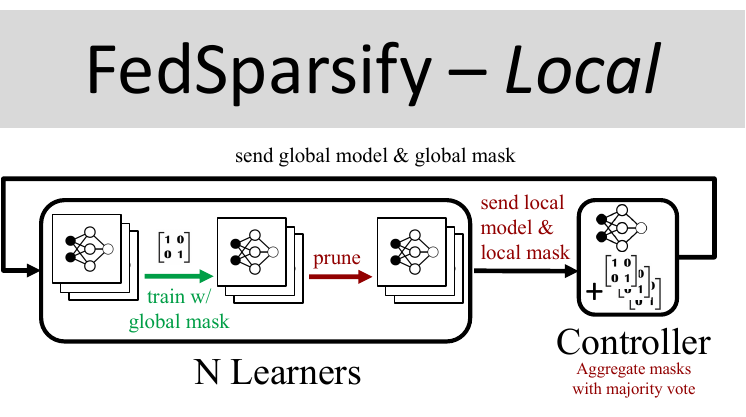}\label{ExecutionFlow_FedSparsifyLocal}
  }
  \subfloat[The model is pruned only by the controller (server). The learners receive the pruned global model and the associated sparsification binary mask. During local training only the non-pruned parameters of the global model are being updated.]{\includegraphics[width=0.35\textwidth]{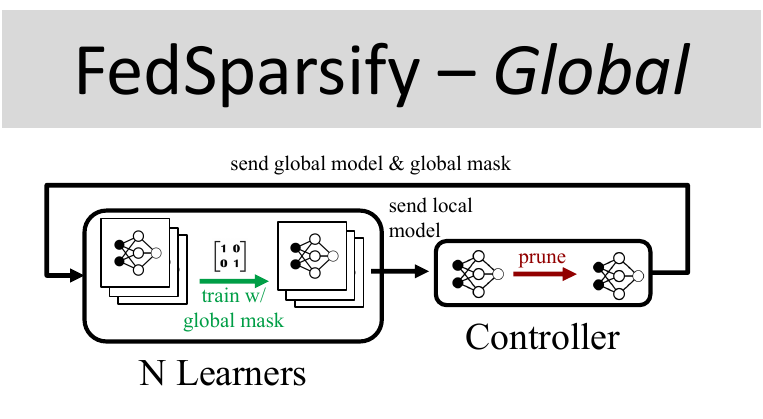}\label{ExecutionFlow_FedSparasifyGlobal}
  }

  \subfloat[Before federated training begins, a participating learner is randomly selected to prune the original model. Once the model is pruned, federated training starts with all learners updating the set of non-pruned parameters enforced by the sparsification mask.] {\includegraphics[width=0.35\linewidth]{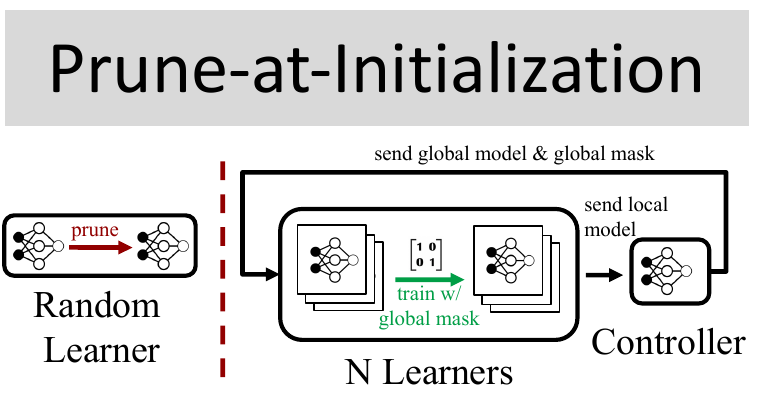}\label{ExecutionFlow_PruneAtInitialization}
  }\qquad
  \subfloat[The global model is pruned right after a specific number of federation rounds is reached. If fine-tuned is enabled (OneShot w/ FineTuning) then the federated model will be fine-tuned (trained) for a couple of rounds.]{\includegraphics[width=0.38\linewidth]{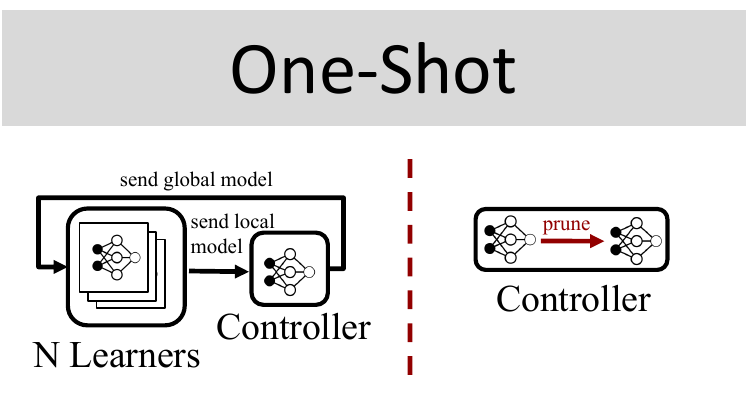}\label{ExecutionFlow_OneShotPruning}
  }
  
  \caption{Execution flow diagrams for different federated sparsification methods.}
  \label{fig:Appendix_SparsificationMethodsExecutionFlow}
\end{figure}

\paragraph{Sparsification Methods Execution Flow.} In Figure~\ref{fig:Appendix_SparsificationMethodsExecutionFlow} we show the execution flow for the two FedSparsify variants, pruning at initialization methods and OneShot pruning.

\begin{figure}[htpb]
  \centering    
  \includegraphics[width=0.7\linewidth]{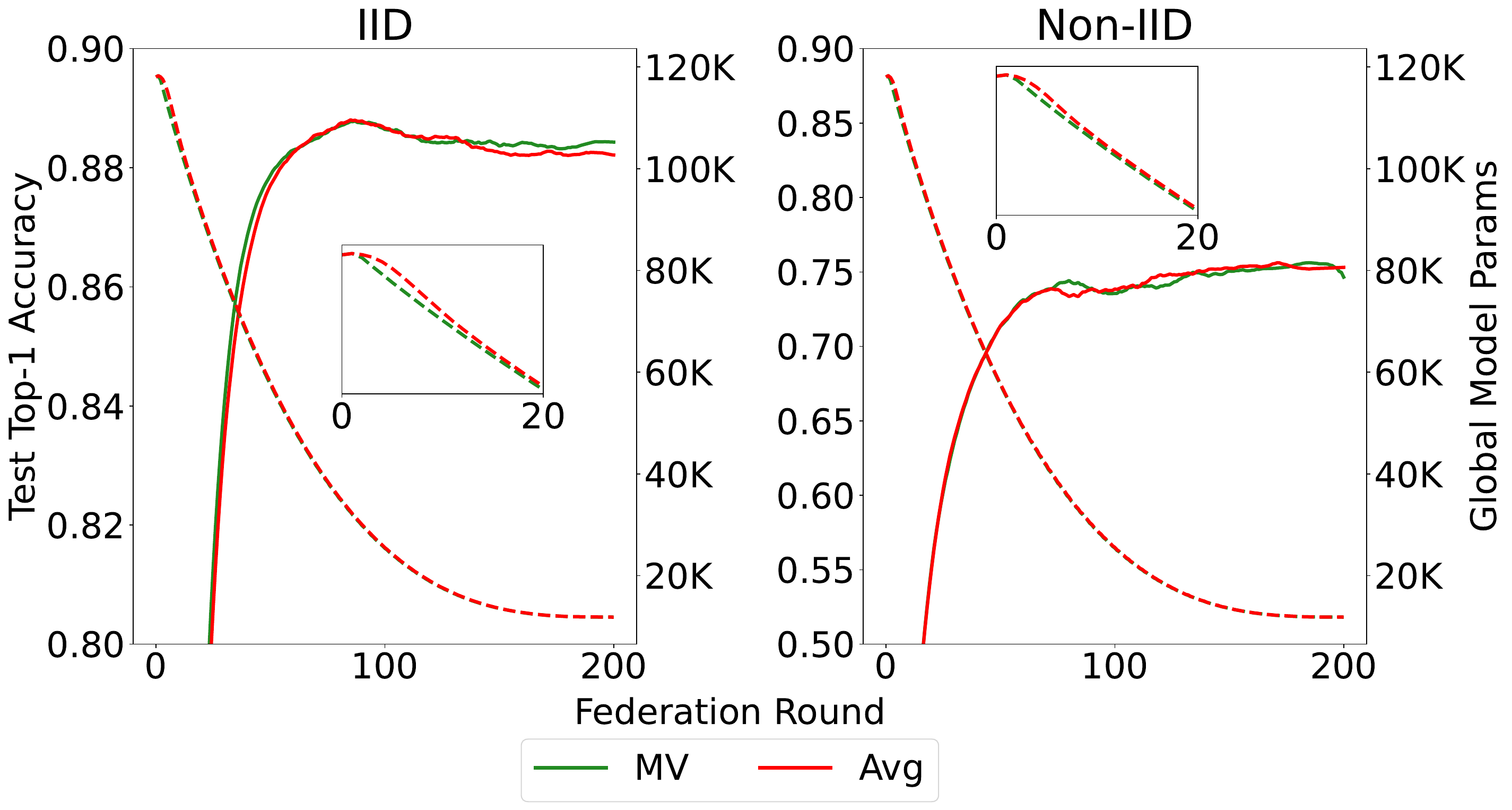}
  \caption{Convergence of FedSparsify-Local with Majority-Voting (MV) as aggregation rule and FedSparsify-Local with Weighted Average (FedAvg/Avg) as aggregation rule on FashionMNIST with 10 clients over IID and Non-IID data distributions at 0.9 sparsity. Left y-axis and solid lines show accuracy, right y-axis, and dashed lines show global model parameters reduction.}
  \label{subfig:Appendix_FashionMNIST_MajorityVsAverageComparison}
\end{figure}

\section{Additional Evaluation}
\label{sec:Appendix_Additional_Evaluation}
\begin{figure}[htpb]
  \centering
  \subfloat[FashionMNIST (FC) - 10 Clients]{
  \includegraphics[width=0.5\linewidth]{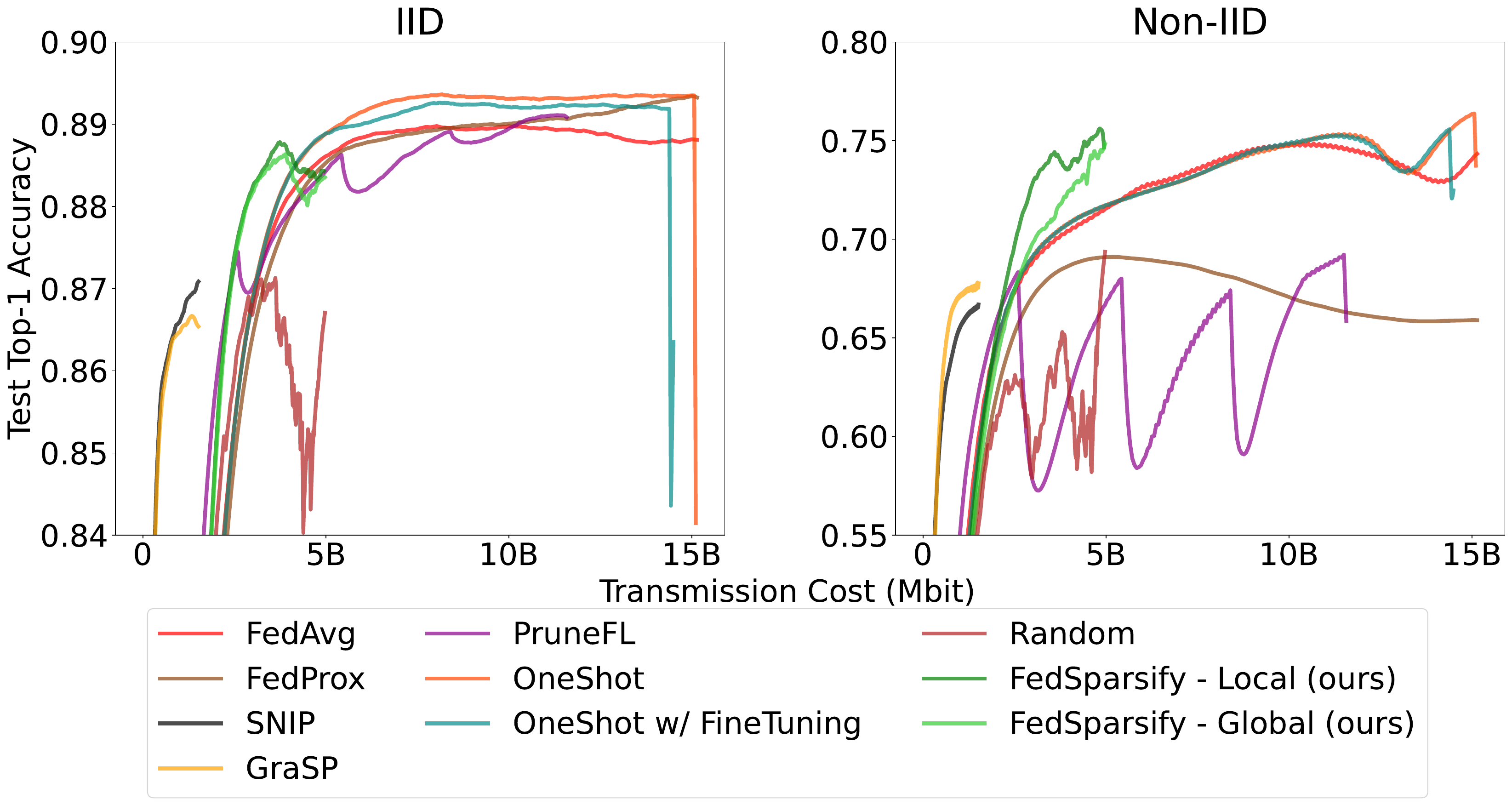}
  \label{subfig:Appendix_FashionMNIST_Convergence_TransmissionCost_10clients}
  }
  \centering
  \subfloat[FashionMNIST (FC) - 100 Clients]{
  \includegraphics[width=0.5\linewidth]{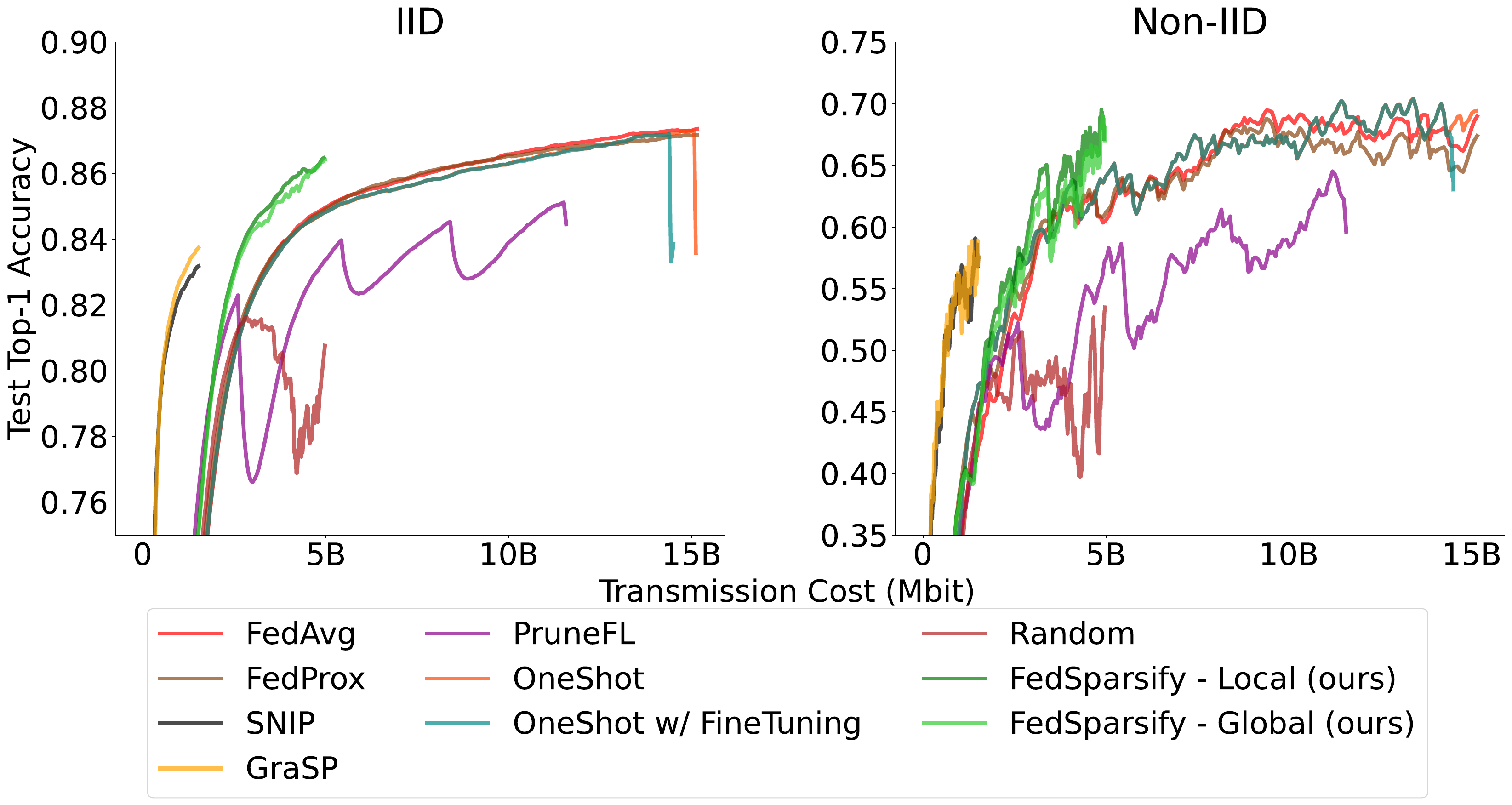}
  \label{subfig:Appendix_FashionMNIST_Convergence_TransmissionCost_100clients}
  }
  
  \subfloat[CIFAR-10 (CNN) - 10 Clients]{
  \includegraphics[width=0.5\linewidth]{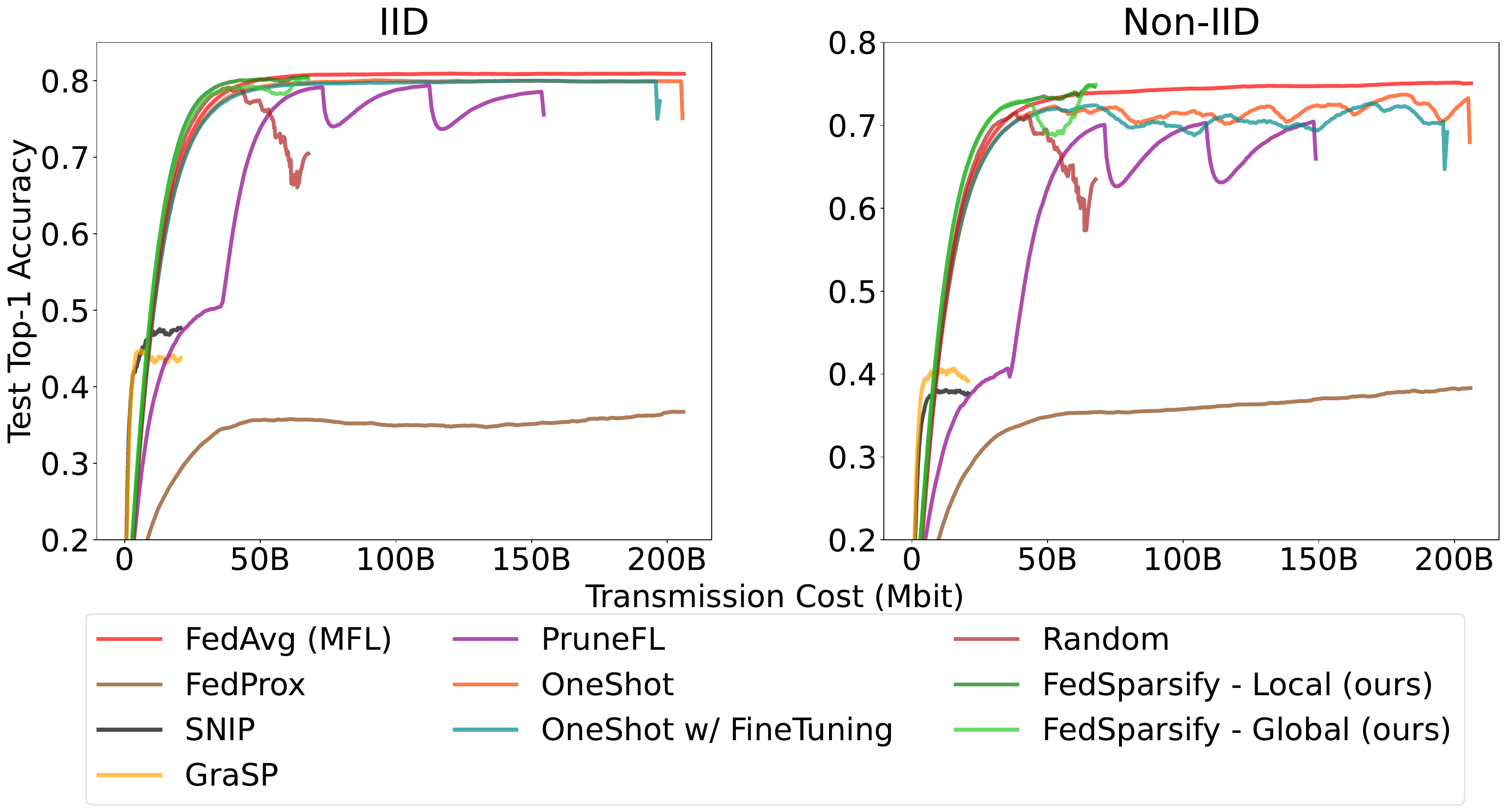}
  \label{subfig:Appendix_Cifar10_Convergence_TransmissionCost_10clients}
  }
  \subfloat[CIFAR-10 (CNN) - 100 Clients]{
  \includegraphics[width=0.5\linewidth]{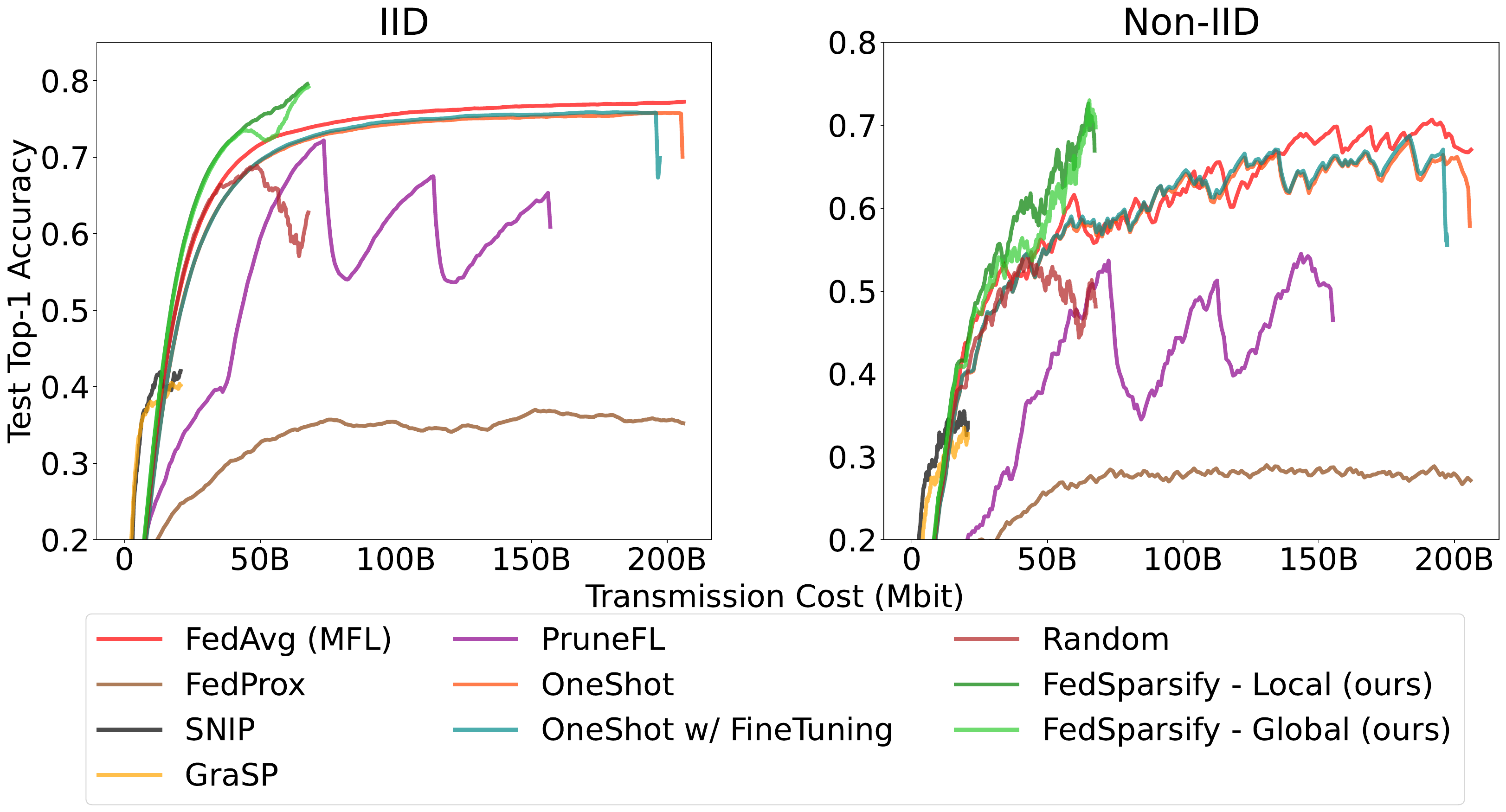}
  \label{subfig:Appendix_Cifar10_Convergence_TransmissionCost_100clients}
  }
  
  \subfloat[CIFAR-100 (VGG) - 10 Clients]{
  \includegraphics[width=0.5\linewidth]{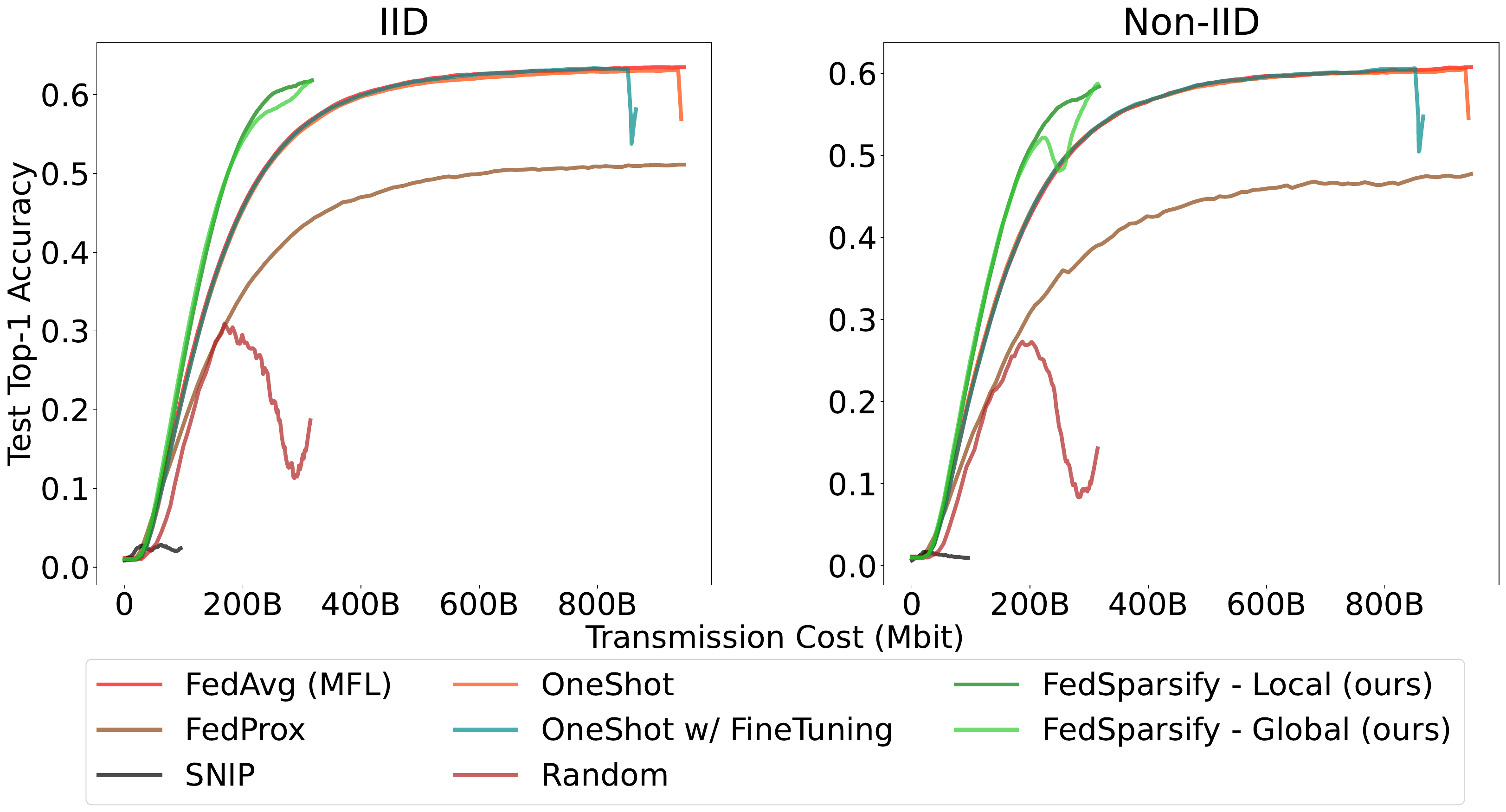}
  \label{subfig:Appendix_Cifar100_Convergence_TransmissionCost_10clients}
  }
  \subfloat[CIFAR-100 (VGG) - 100 Clients]{
  \includegraphics[width=0.5\linewidth]{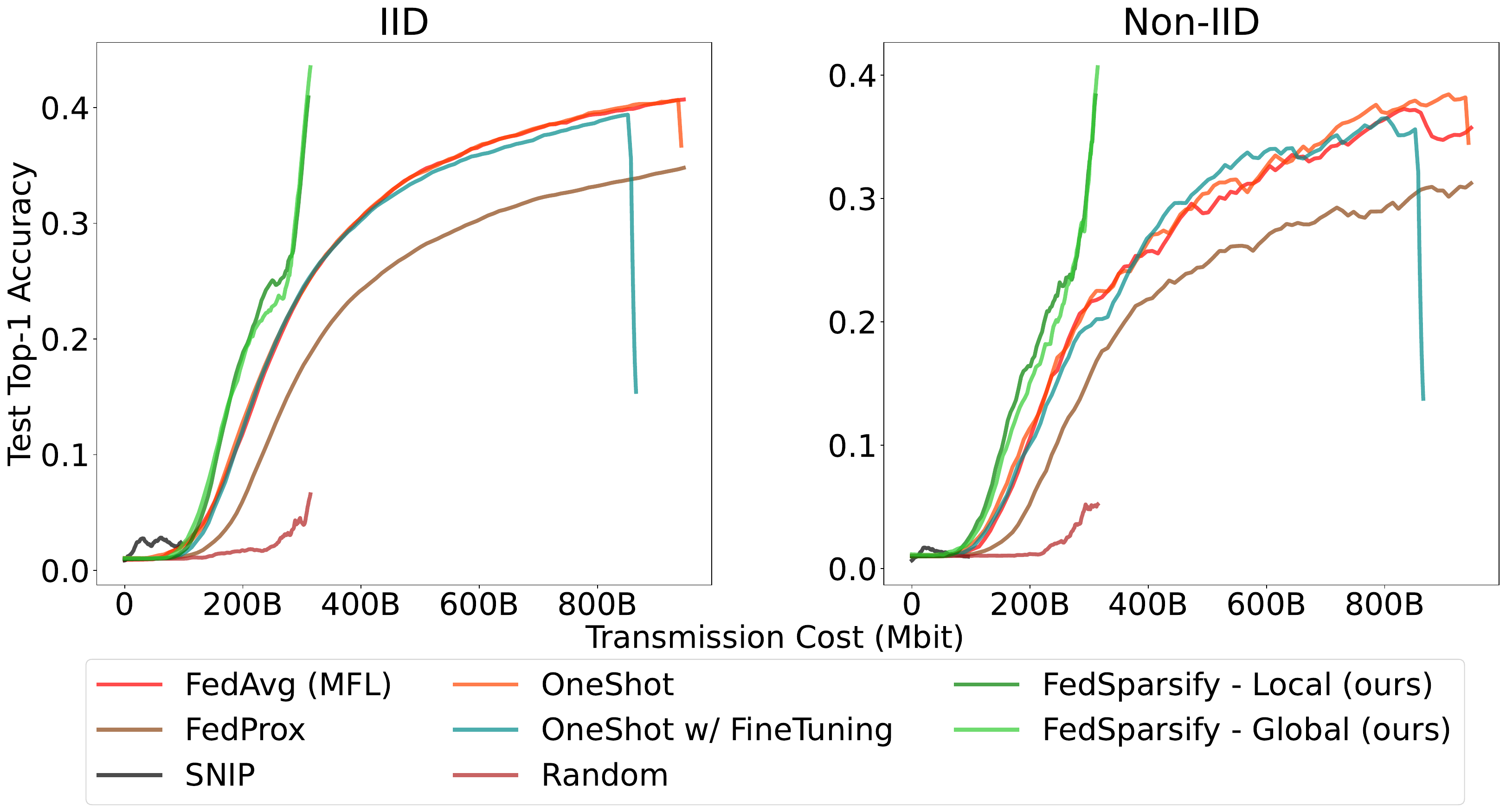}
  \label{subfig:Appendix_Cifar100_Convergence_TransmissionCost_100clients}
  }
  
  \caption{Transmission Cost vs. Accuracy for FashionMNIST (top row), CIFAR-10 (middle row) over the course of 200 federation rounds and for CIFAR-100 (bottom row) over the course of 100 federation rounds. Across all environments, SNIP, GraSP, Random, FedSparsify-Local and FedSparsify-Global convergence is shown at 0.9 model sparsity and PruneFL at 0.3.}
  \label{fig:Appendix_Convergence_TransmissionCost}
\end{figure}

In this section, we highlight the difference between Majority Voting and FedAvg as aggregation rules for \textit{FedSparsify-Local} and we provide a detailed description of federated models convergence in terms of cumulative transmission (communication) cost.
Figure~\ref{fig:Appendix_Convergence_TransmissionCost} shows this convergence across all four federated environments, i.e., 10 and 100 clients at IID and Non-IID environments with a sparsification rate of 0.9 for all sparsification schemes except for PruneFL, which is shown at 0.3 (recommended sparsity). In Table~\ref{tbl:Appendix_FashionMNIST_ModelsComparison}, we show a thorough quantitative comparison of sparse and non-sparse federated models' throughput, inference, size and communication cost for the FashionMNIST dataset in the environment of 10 learners. We used the DeepSparse library (\url{https://github.com/neuralmagic/deepsparse}) to measure inference times. For all models, we stress-test their inference time by allocating a total execution time of 60 seconds with a warmup period of 10 seconds.

\paragraph{Majority Voting Aggregation.} In Figure~\ref{subfig:Appendix_FashionMNIST_MajorityVsAverageComparison}, we show the learning performance (left y-axis) and global model parameters decrease (right y-axis) for the federated FashionMNIST model in a federated environment of 10 clients trained using the \textit{FedSparsify-Local} sparsification schedule when Majority Voting and FedAvg are used as the aggregation rule of learners' local models. As it is shown (inset of the figure) at the beginning of training Majority Voting preserves the sparsity of the local models enforced by clients' local masks, while FedAvg prunes lesser parameters.

\paragraph{Transmission Cost.} We measure transmission costs in terms of Megabits (Mbit) exchanged for all federated training rounds. We visualize transmission cost during training in Figure~\ref{fig:Appendix_Convergence_TransmissionCost}. We train models for a fixed number of rounds, therefore all models do not have the same transmission cost at the end of training. We compute the cost of transmitting parameters without any compression, i.e., the transmission cost at each round is the total number of clients participating at each round, multiplied by the total number of non-zero parameters sent by the server at the beginning of the round (i.e., global model size) to all participating clients, plus the total number of non-zero parameters uploaded to the server by all participating clients at the end of the round. We multiply this aggregated quantity by 32, assuming full-precision training. If the sparsification scheme exchanges binary masks with the server during federated training (e.g., \textit{FedSparsify-Local}) then we also add to this quantity the total number of parameters of the original model, i.e., the size of the binary mask (1-bit parameters) is equal to the original model size without any sparsification. 

\begin{figure}[htpb]
  \centering
  \subfloat[FashionMNIST (FC) - 10 Clients]{
  \includegraphics[width=0.5\linewidth]{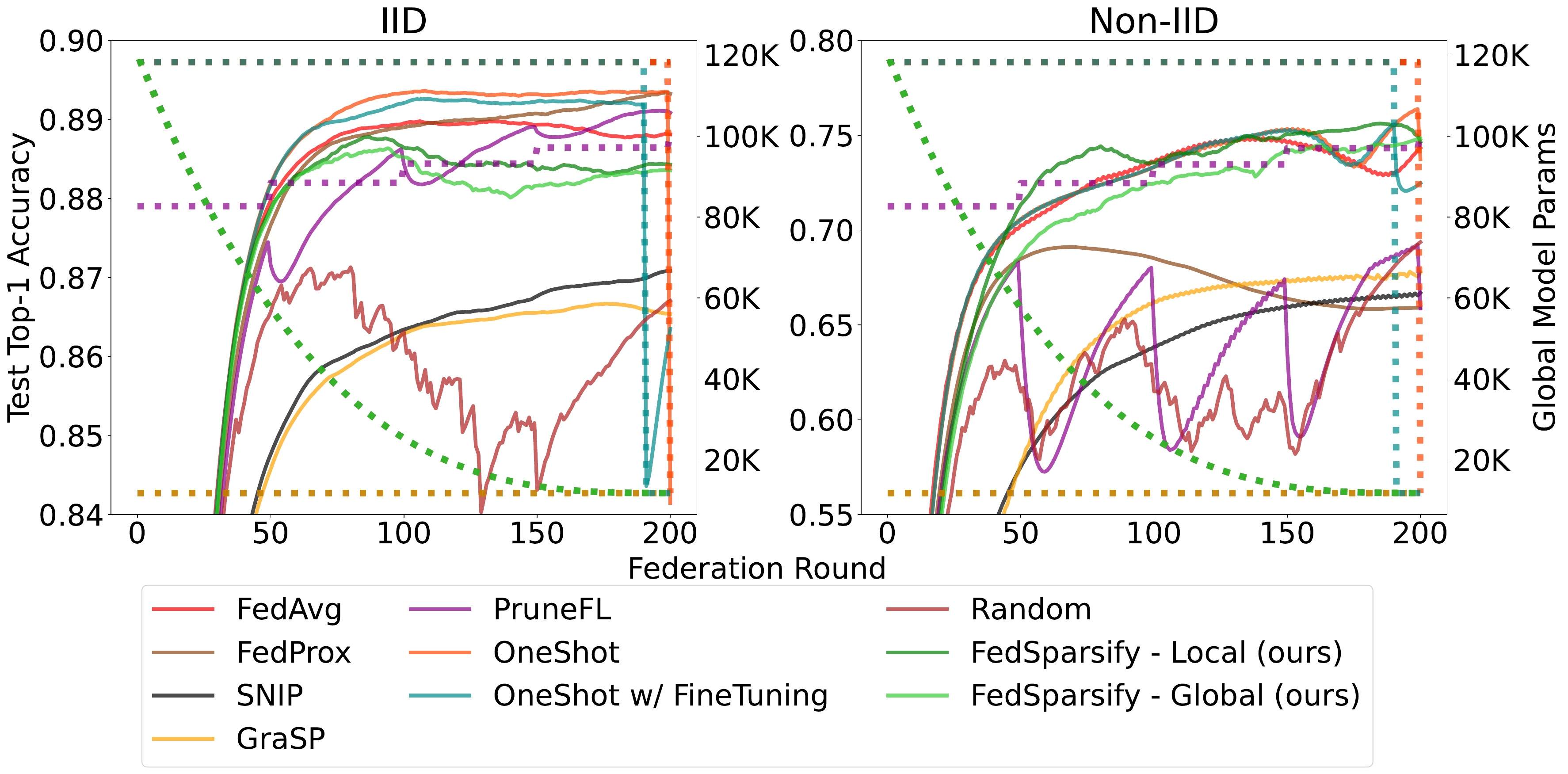}
  \label{subfig:Appendix_FashionMNIST_Convergence_FederationRoundsGlobalModelParams_Comparison_10clients}
  }
  \centering
  \subfloat[FashionMNIST (FC) - 100 Clients]{
  \includegraphics[width=0.5\linewidth]{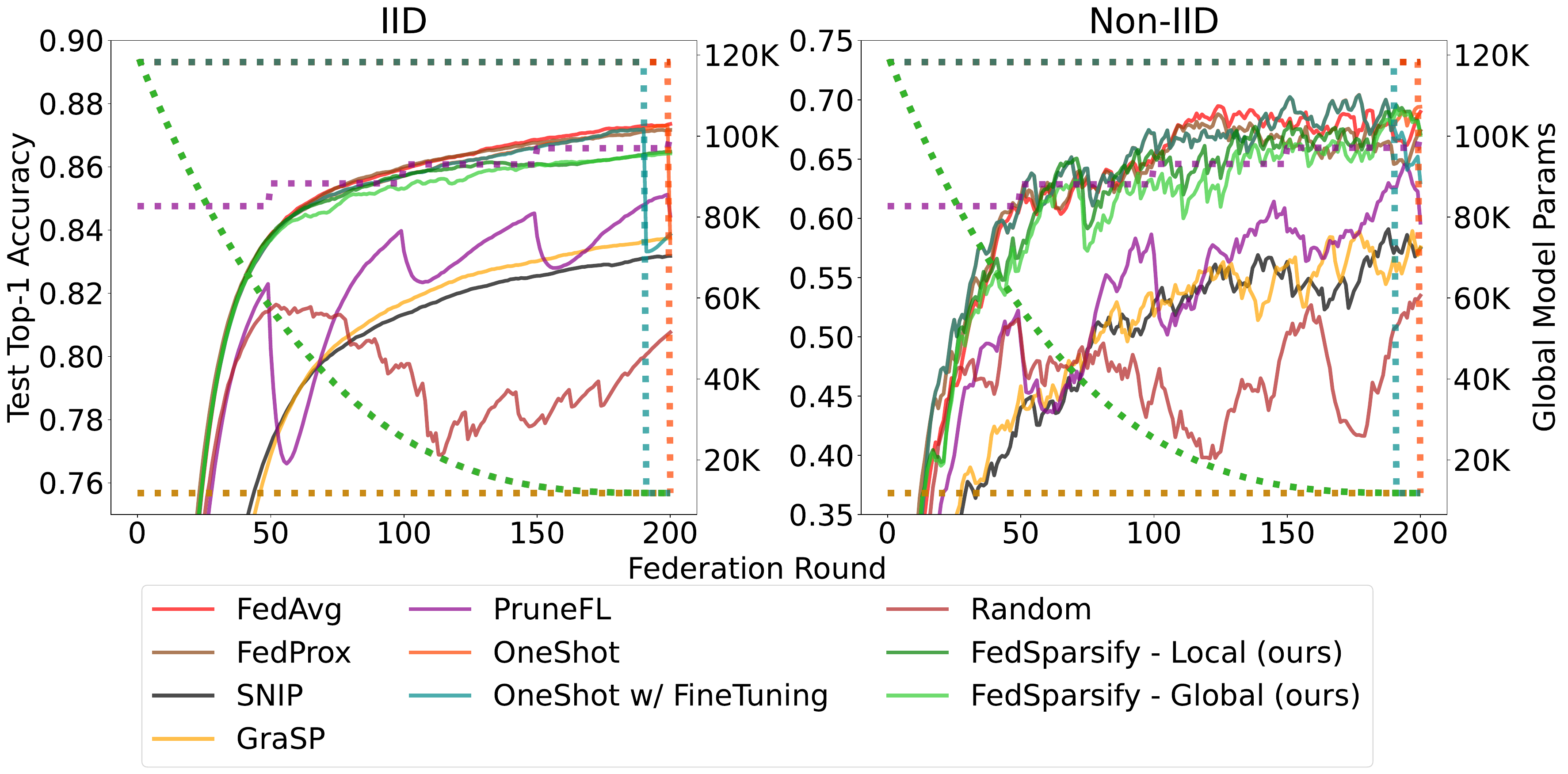}
  \label{subfig:Appendix_FashionMNIST_Convergence_FederationRoundsGlobalModelParams_Comparison_100clients}
  }
  
  \subfloat[CIFAR-10 (CNN) - 10 Clients]{
  \includegraphics[width=0.5\linewidth]{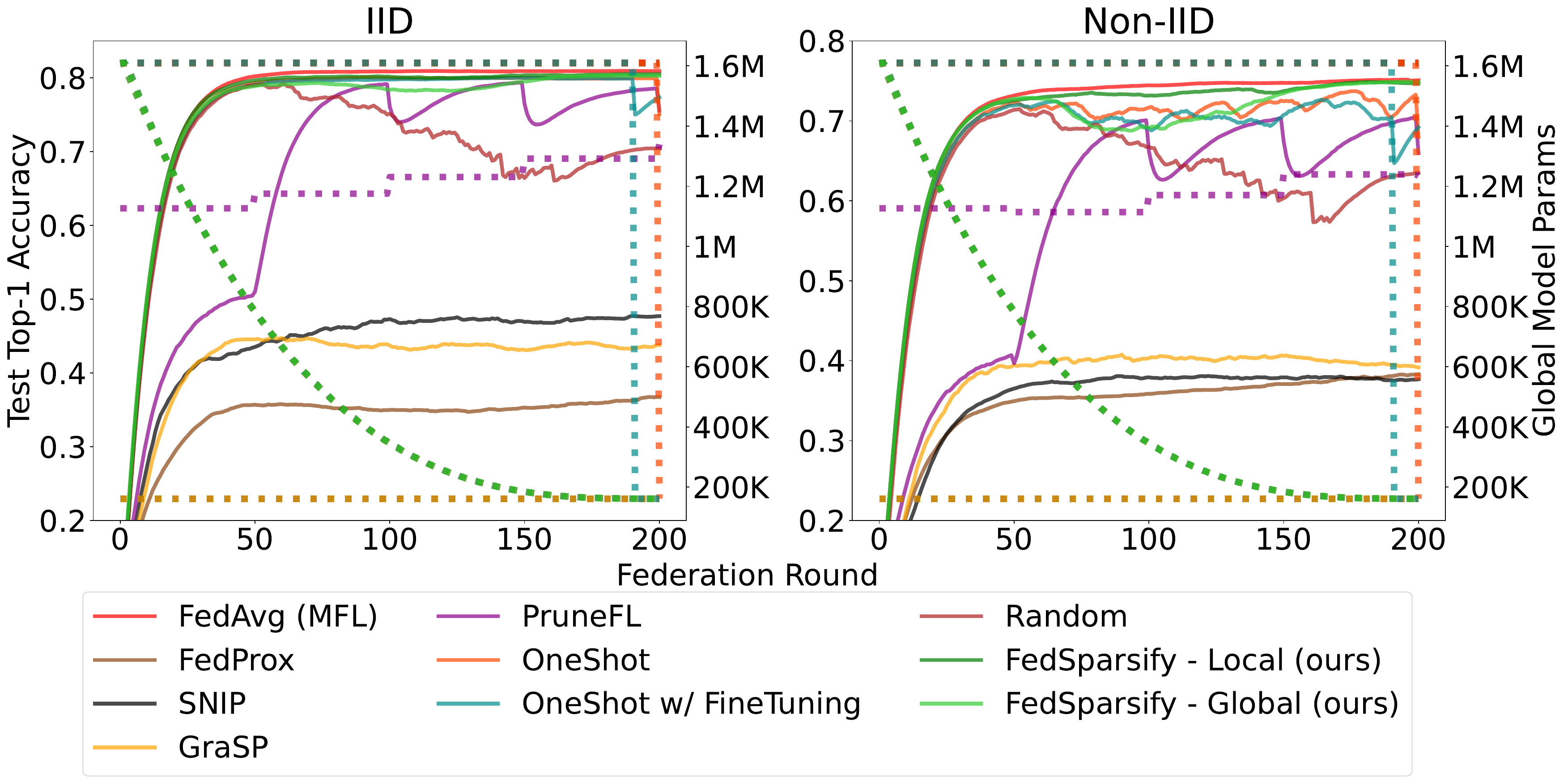}
  \label{subfig:Appendix_Cifar10_Convergence_FederationRoundsGlobalModelParams_Comparison_10clients}
  }
  \subfloat[CIFAR-10 (CNN) - 100 Clients]{
  \includegraphics[width=0.5\linewidth]{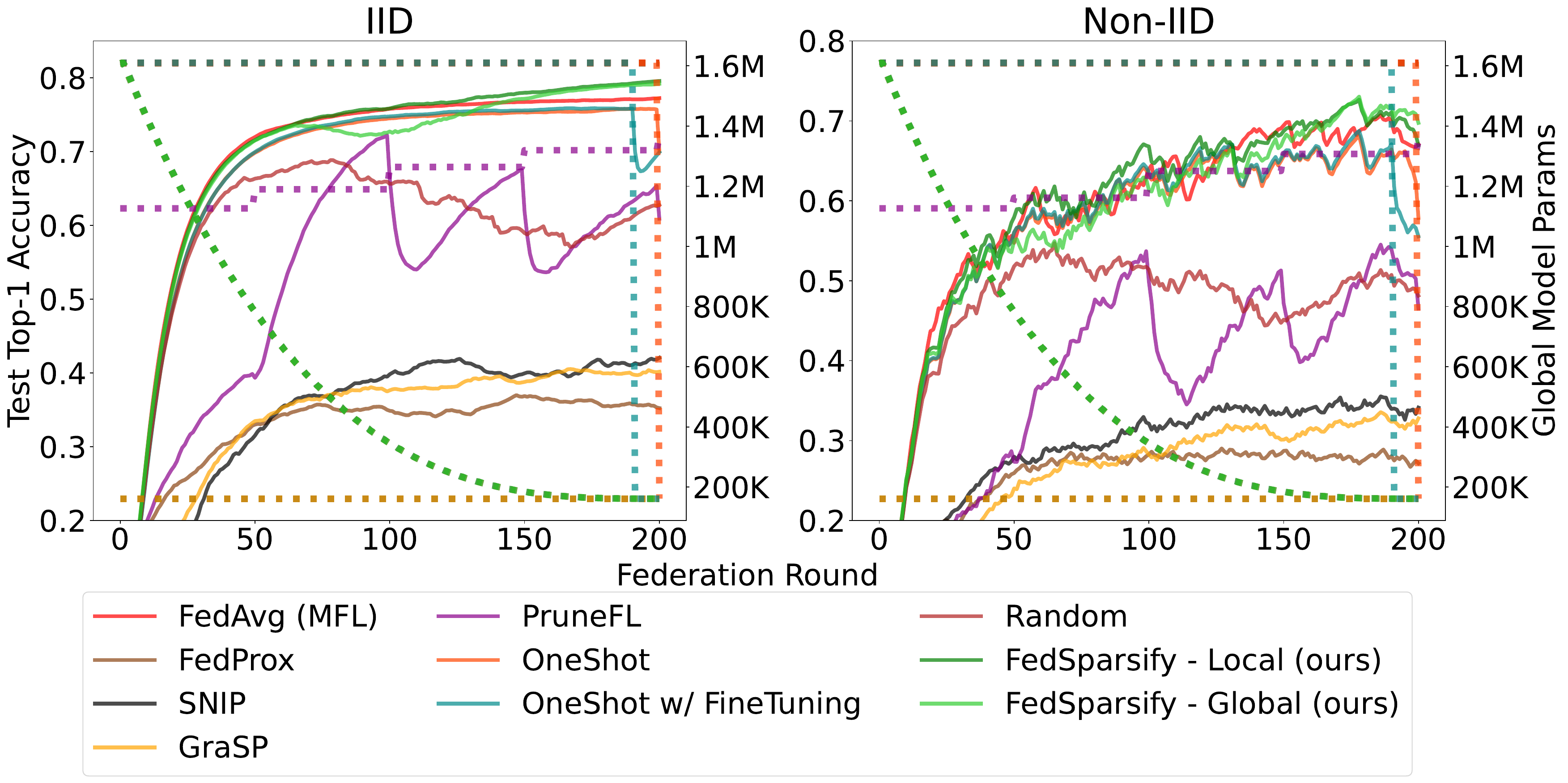}
  \label{subfig:Appendix_Cifar10_Convergence_FederationRoundsGlobalModelParams_Comparison_100clients}
  }
  
  \subfloat[CIFAR-100 (VGG) - 10 Clients]{
  \includegraphics[width=0.5\linewidth]{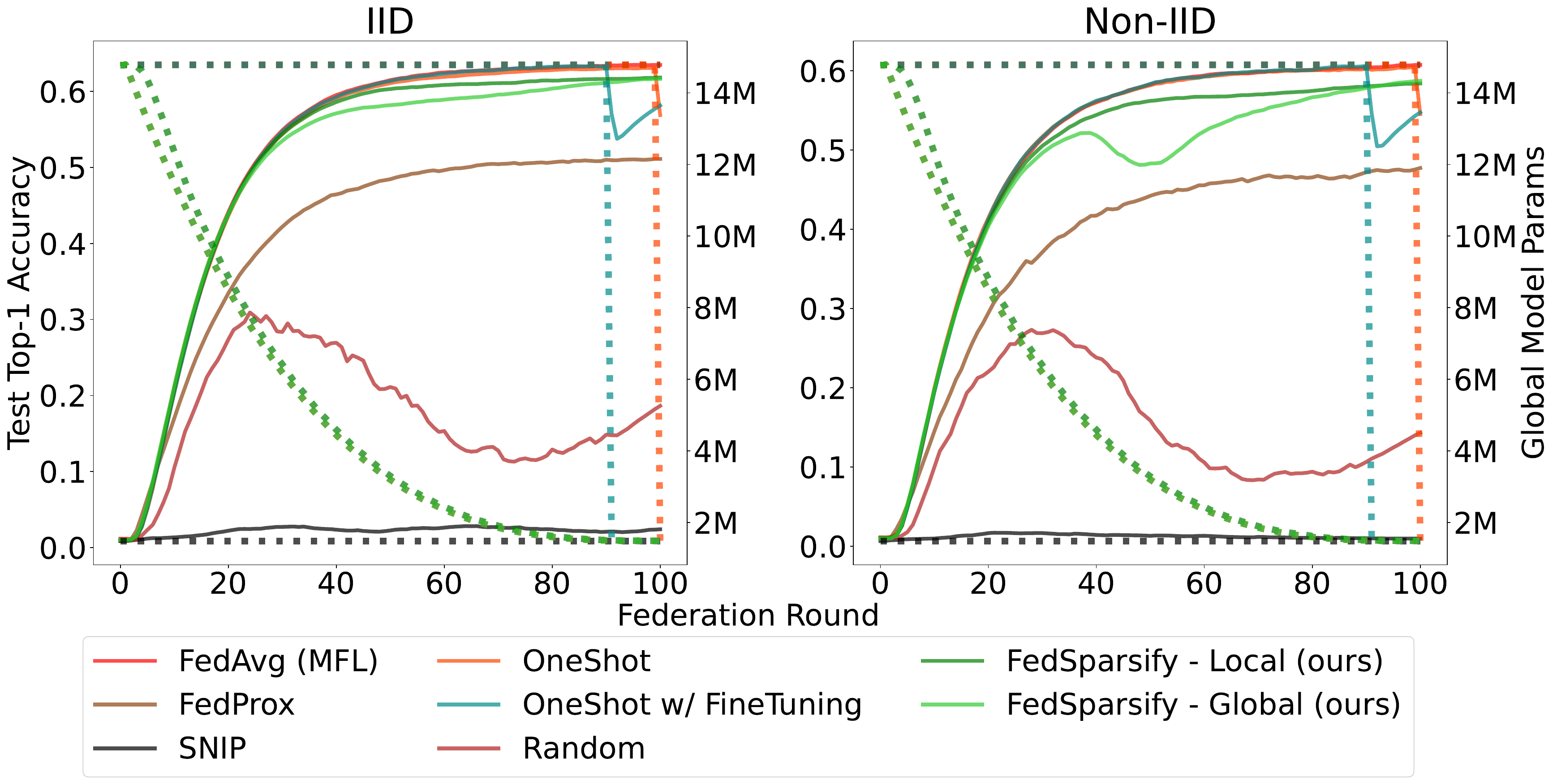}
  \label{subfig:Appendix_Cifar100_Convergence_FederationRoundsGlobalModelParams_Comparison_10clients}
  }
  \subfloat[CIFAR-100 (VGG) - 100 Clients]{
  \includegraphics[width=0.5\linewidth]{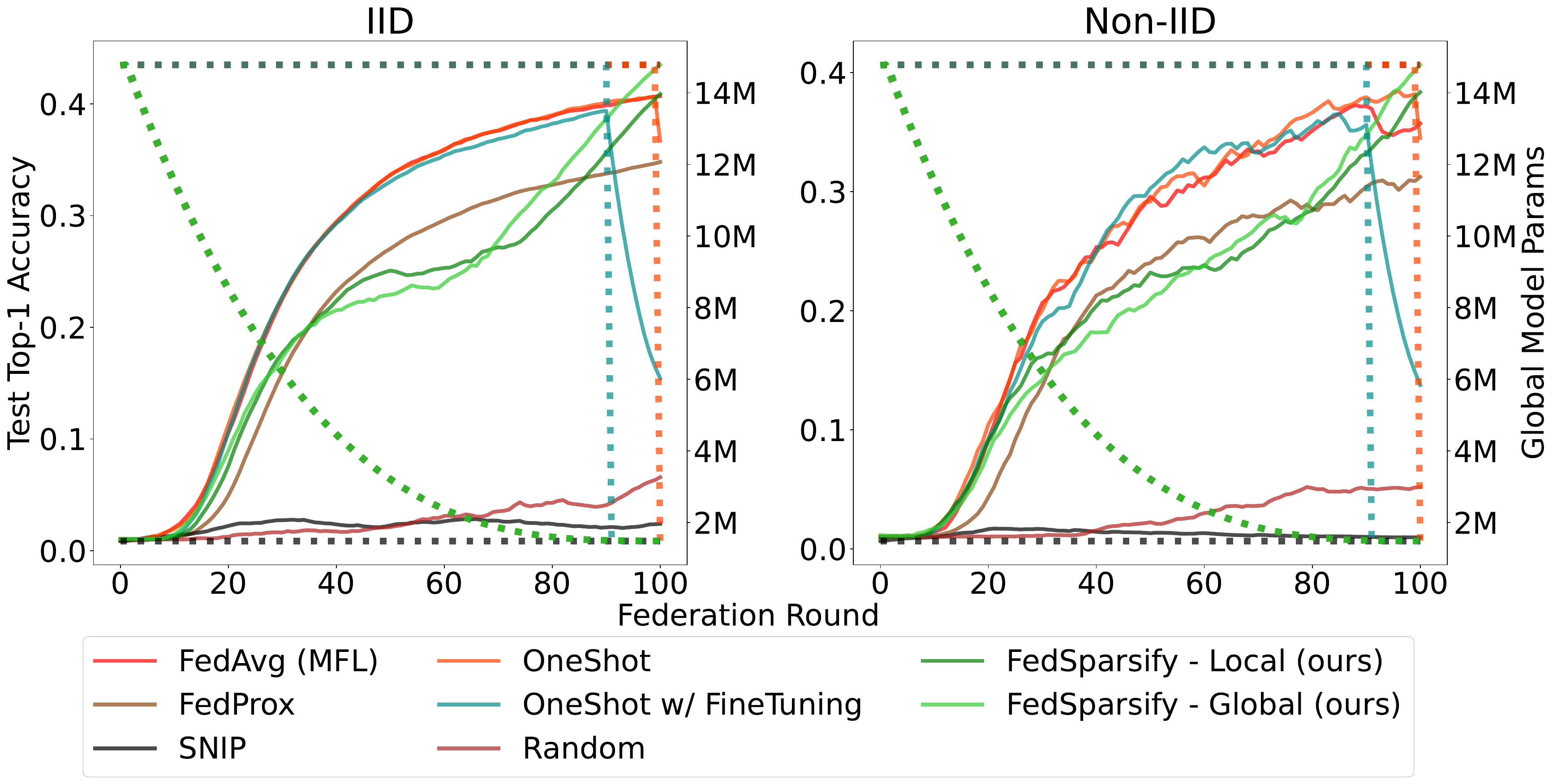}
  \label{subfig:Appendix_Cifar100_Convergence_FederationRoundsGlobalModelParams_Comparison_100clients}
  }
  
  \caption{Transmission Cost vs. Accuracy for FashionMNIST (top row), CIFAR-10 (middle row) over the course of 200 federation rounds and for CIFAR-100 (bottom row) over the course of 100 federation rounds. Across all environments, SNIP, GraSP, Random, FedSparsify-Local and FedSparsify-Global convergence is shown at 0.9 model sparsity and PruneFL at 0.3.}
  \label{fig:Appendix_Convergence_FederationRound}
\end{figure}

As shown in Figures~\ref{subfig:Appendix_FashionMNIST_Convergence_TransmissionCost_10clients}, \ref{subfig:Appendix_Cifar10_Convergence_TransmissionCost_10clients}, and~\ref{subfig:Appendix_Cifar100_Convergence_TransmissionCost_10clients} for FashionMNIST, CIFAR-10 and CIFAR-100 for the same number of Mbits exchanged between the clients and the server, FedSparsify achieves a higher learning performance when compared to other no-pruning (FedAvg) and pruning baselines (PruneFL). On the contrary, though, pruned federated models learned through SNIP and GraSP schemes exchange significantly fewer model parameters compared to the rest of the schemes, but they require more synchronization rounds to perform at par with other schemes in FashionMNIST. In more challenging setups (cf. CIFAR-10) SNIP and GraSP do not outperform other approaches or fail to learn (cf. CIFAR-100) even with a small number of transmitted Mbits (e.g., 30k in CIFAR-10, 100k in CIFAR-100). Across all environments with 100 clients, see Figures~\ref{subfig:Appendix_FashionMNIST_Convergence_TransmissionCost_100clients},~\ref{subfig:Appendix_Cifar10_Convergence_TransmissionCost_100clients} and~\ref{subfig:Appendix_Cifar100_Convergence_TransmissionCost_100clients}, FedSparsify successfully learns a highly performant model that greatly outperforms all other approaches for the same number of exchanged Mbits (e.g., 5k Mbit in FashionMNIST, 50k Mbit in CIFAR-10, 400k Mbit in CIFAR-100). Interestingly, in the case of CIFAR-100 FedSparsify exhibits a learning spike towards the end of training which is also evident in CIFAR-10 and Non-IID. The spike is attributed to the small number of model parameters exchanged during FedSparsify training, since towards the end of training the global model size is very close to 0.9 sparsity (cf. Figures~\ref{subfig:Appendix_Cifar10_Convergence_FederationRoundsGlobalModelParams_Comparison_100clients},~\ref{subfig:Appendix_Cifar100_Convergence_FederationRoundsGlobalModelParams_Comparison_100clients}).

\begin{table*}[htpb]
    \centering
    \small
    \centering
    \setlength\tabcolsep{1.5pt}
    \begin{tabular}{@{}cccccccc@{}}
        \toprule
        Sparsity & Accuracy & Params & Model Size (MBs) & C.C. (MM) & Inf.Latency & Inf.Iterations & Inf.Throughput \\ \midrule
        0.0 & 0.7489 & 118,282 & 0.434 & 473 & 0.607 & 755,817 & 403,096 \\
        0.8 & 0.74 & 23,657 & 0.109 (x3.97) & 190 (x2.48) & 0.601 (x1.01) & 763,298 (x1.01) & 407,085 (x1.01) \\
        0.85 & 0.735 & 17,743 & 0.082 (x5.24) & 173 (x2.73) & 0.594 (x1.02) & 772,976 (x1.02) & 412,251 (x1.02) \\
        0.90 & 0.749 & 11,829 & 0.056 (x7.75) & 155 (x3.04) & 0.588 (x1.03) & 781,005 (x1.03) & 416,532 (x1.03) \\
        0.95 & 0.735 & 5,915 & 0.029 (x14.68) & 137 (x3.43) & 0.587 (x1.03) & 783,000 (x1.03) & 417,596 (x1.03) \\
        0.99 & 0.687 & 1,183 & 0.008 (x53.95) & 123 (x3.82) & 0.58 (x1.04) & 792,332 (x1.04) & 422,569 (x1.04) \\
        \bottomrule
    \end{tabular}
    \caption{Federated models comparison for FashionMNIST in the \textit{Non-IID} environment of 10 clients. All recorded values are measurements from the model learned at the end of federated training for a total number of 200 federation rounds. All sparsified models represent the execution results of FedSparsify-Global and sparsity 0.0 of FedAvg. C.C. is an abbreviation for communication cost in terms of the total number of exchanged parameters, expressed in millions (MM). Model inference efficiency is measured by mean processing time per batch (Inf.Latency - ms/batch), the number of iterations (Inf.Iterations), and processed items per second (Inf.Throughput - items/sec). Values in parenthesis show x-times reduction (for model size, communication cost and inference latency) and x-times increase/speedup (for inference iterations and throughput) compared to non-pruning.}
    \label{tbl:Appendix_FashionMNIST_ModelsComparison}
\end{table*}

\end{document}